\documentclass[10pt,twocolumn,letterpaper]{article}

\usepackage[pagenumbers]{wacv} 

\usepackage{threeparttable}
\usepackage{multirow}
\usepackage{tabularx}

\definecolor{wacvblue}{rgb}{0.21,0.49,0.74}
\usepackage[pagebackref,breaklinks,colorlinks,allcolors=wacvblue]{hyperref}

\def\wacvPaperID{726} 
\def\confName{WACV}
\def\confYear{2026}

\title{Dense Coverage, Sparse Refinement: Byte-Constrained Cooperative Perception}

\author{%
  Melih Yazgan\textsuperscript{1,2}\textsuperscript{†}\quad
  Timon Mueller\textsuperscript{1}\textsuperscript{*}\quad
  J.~Marius Zoellner\textsuperscript{1,2}\\
  \textsuperscript{1}FZI Research Center for Information Technology, 
  \textsuperscript{2}Karlsruhe Institute of Technology\\
  \footnotesize\texttt{last.name@fzi.de}
}

\begin{document}
\maketitle
\begin{abstract}
Collaborative perception improves autonomous perception by sharing intermediate Bird’s-Eye-View (BEV) features across connected agents, but dense feature exchange is difficult to deploy under strict Vehicle-to-Everything (V2X) bandwidth limits. Existing efficient methods typically either compress the full feature map uniformly, spending bits on low-value background, or sparsify communication, risking the loss of useful context. We propose a coverage-refinement design for byte-constrained cooperative perception: each agent transmits a highly compressed coarse layer over the full BEV map and allocates the remaining budget to selected high-resolution patches. A Task-Aware Benefit Selector ranks cells by estimated downstream utility, enabling deterministic budgeted refinement and zero-retraining adaptation to changing bandwidth. The receiver reconstructs a dense BEV tensor compatible with standard fusion modules. Experiments on DAIR-V2X and OPV2V show strong accuracy--payload
trade-offs at kilobyte-scale budgets. On DAIR-V2X, our method reaches
0.60 AP@0.7 at only 1.87 KB per non-ego agent, compared with 0.52 at
4.61 KB for uniform SimVQ compression. Controlled diagnostics further
show that the gain arises from coverage--refinement allocation rather
than quantization alone. Code will be published.
\end{abstract}    
\section{Introduction}

Collaborative perception (CP) extends the sensing range of autonomous agents by sharing intermediate Bird's-Eye-View (BEV) features across vehicles and infrastructure~\cite{wang_v2vnet_2020, xu_cobevt_2022, sadafusion, Yazgan_2026_CVPR}. However, dense BEV tensors are expensive to transmit and can exceed practical Vehicle-to-Everything (V2X) bandwidth. This creates a central challenge: how can agents preserve the benefit of dense collaborative features while communicating only a few kilobytes?

Existing communication-efficient CP methods mainly follow three paradigms. Global compression methods, such as autoencoders~\cite{hmvit, xu2022v2xvit, core} or vector quantization~\cite{revqom2025}, preserve dense spatial support but allocate bits uniformly, including to task-irrelevant background. Spatial sparsification methods~\cite{hu_where2comm_2022, how2comm, Yazgan_2025_ITSC, infocom, select2col, scope} transmit only selected high-utility regions, but their select-or-discard design removes unselected context and may require missingness-aware fusion. Interactive query-based methods~\cite{cocmt, instinct, xu_cosdh_2025} can reduce redundancy further, but multi-round handshakes introduce latency and are less suitable for one-shot V2X broadcast~\cite{raca2020beyond}.

We argue that byte-constrained CP should be formulated as a \emph{coverage-refinement} problem. Instead of choosing between dense uniform compression and sparse incomplete transmission, each agent should first provide a coarse but dense representation of the full scene, and then spend the remaining bytes on fine details only where they are most useful for detection. Based on this principle, we propose a dual-resolution communication strategy. A heavily compressed coarse base layer guarantees dense BEV coverage and preserves compatibility with standard fusion modules. A Task-Aware Benefit Selector then estimates the marginal utility of refining each spatial cell and selects the highest-benefit cells for fine transmission under a strict payload budget. Since fine cells have constant bit cost, inference reduces to deterministic ranked-prefix selection, enabling instant budget adaptation without retraining or multi-round communication.

Our contributions are threefold:
\begin{itemize}
    \item \textbf{Coverage-refinement communication.}
    We reformulate bandwidth-limited cooperative perception as dense coarse coverage plus sparse fine refinement, preserving a standard dense BEV fusion interface while avoiding the complete information loss of sparse-only communication.

    \item \textbf{Task-aware budget allocation.}
    We introduce a sender-side Task-Aware Benefit Selector that ranks spatial cells by the expected downstream value of fine transmission, yielding deterministic, nested patch selections across changing byte budgets.

    \item \textbf{Controlled validation of the allocation principle.}
Through frozen-codec ranking controls, no-quantization diagnostics,
no-coarse ablations, and dynamic-budget evaluation, we show that the
gain stems from task-aware coverage--refinement allocation rather than
from quantization alone, while preserving zero-retraining adaptation.
\end{itemize}
\section{Related Work}
Communication efficiency is a central challenge in intermediate-fusion cooperative perception, with prior methods spanning compression, selective communication, and combined strategies~\cite{yazgan_survey_2024}. We organize the most relevant recent methods into three practical communication paradigms: interactive/query-based coordination, sender-side spatial sparsification, and global feature compression. We also discuss layered, scalable coding as a related design principle for our coverage-refinement formulation.

\noindent \textbf{Interactive and Query-Based Collaboration.}
Interactive methods~\cite{liu_who2com_2020, hu_communication-efficient_2024, Yazgan_2025_ICCV}, including CoSDH~\cite{xu_cosdh_2025}, JigsawComm$^\dagger$~\cite{jigsaw}, and WhisperNet$^\dagger$~\cite{chen2026whispernet}, reduce redundancy by letting the receiver or ego agent coordinate complementary information from collaborators. CoSDH follows a supply--demand request--response design, while WhisperNet first exchanges lightweight spatial--channel saliency metadata and then allocates communication budgets across agents, spatial regions, and feature channels. Such query-driven or receiver-coordinated designs can be highly bandwidth-efficient because the ego agent explicitly requests missing, uncertain, or complementary information. However, they typically require multi-stage communication, metadata exchange, or score-map exchange, which introduces additional latency and makes deployment more difficult in one-shot, connectionless V2X broadcast settings.

\noindent \textbf{Spatial Sparsification and Filtering.}
Sender-side sparse communication avoids multi-round exchange by transmitting only selected regions. Where2comm~\cite{hu_where2comm_2022} masks low-confidence BEV regions using spatial confidence maps, whereas EffiComm~\cite{Yazgan_2025_ITSC} further adapts per-agent sparsity through GNN-based grid reduction and MoE attention fusion. Task-oriented and information-bottleneck methods such as InfoCom~\cite{infocom}, PragComm$^\dagger$~\cite{pragcomm}, and RDComm$^\dagger$~\cite{RDComm}\footnote{$^\dagger$Lacks official open-source code for standardized benchmark evaluation.} aim to transmit compact task-relevant representations. These approaches are effective under limited bandwidth, but their select-or-discard design removes unselected regions entirely. Under severe byte budgets, this can discard weak but useful context and may require the receiver or fusion module to explicitly handle missing regions using masks, sparse attention, or other mechanisms that account for missingness.

\noindent \textbf{Global Feature Compression.}
A complementary line of work compresses the full feature tensor using autoencoders, channel reduction, entropy coding, or vector quantization (VQ)~\cite{hmvit, xu2022v2xvit, core, revqom2025, quantv2x}. These methods preserve dense spatial support and are therefore compatible with standard dense fusion modules. However, uniform compression allocates capacity across the entire BEV map, regardless of downstream task utility. As a result, scarce bits may be spent on static background or low-value regions, while foreground objects and localization-sensitive structures receive the same resolution as irrelevant areas.

\noindent \textbf{Layered and Scalable Communication.}
Layered base-enhancement coding is a classical strategy for scalable transmission~\cite{schwarz2007overview, ohm2005advances}. In cooperative perception, the same principle is attractive because a receiver often expects a dense BEV tensor for fusion. However, the objective differs from conventional reconstruction: the transmitted representation should maximize downstream detection accuracy under a strict byte budget. Our method instantiates this idea in a task-aware form, using a coarse base layer for dense coverage and a learned refinement layer for high-utility regions.

\noindent \textbf{Positioning.}
Our approach bridges global compression and spatial sparsification. Unlike uniform VQ, it allocates bits according to task utility; unlike sparse-only communication, it preserves dense scene coverage through a coarse base layer. Sparse fine refinements are then added under a strict byte budget, yielding a standard dense BEV tensor without modifying downstream fusion.


\section{Methodology}
\label{sec:method}

\begin{figure*}[t]
  \centering
  \includegraphics[width=\linewidth]{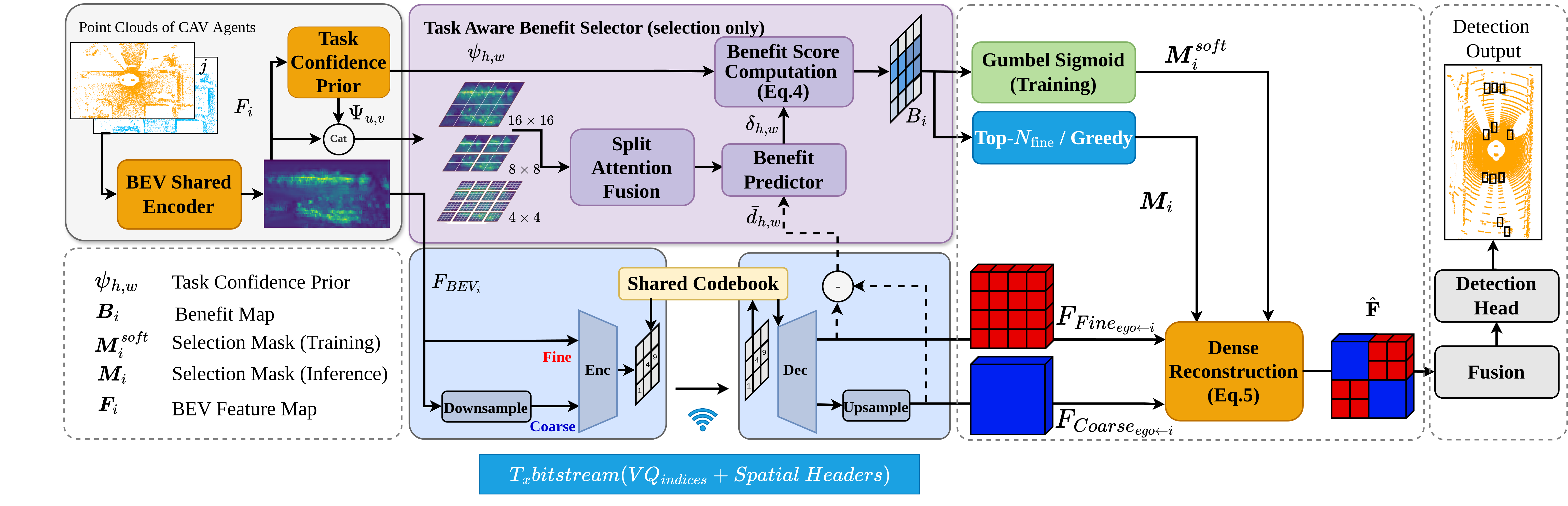}
\caption{\textbf{Overview of our Dual-Resolution Budget-Aware Transmission Framework.} CAV $i$ extracts BEV features $\mathbf{F}_i$ (\cref{sec:met_feature}) and computes a pooled task-confidence prior $\psi_{h,w}$ (\cref{sec:met_task_aware}). Guided by $\psi_{h,w}$, the selector predicts a benefit map $B_i$ to output a soft mask $M_i^{\text{soft}}$ for budgeted training via dual ascent (\cref{sec:differentiable_selection,sec:met_dual_asc}), or a deterministic Top-\(N_{\text{fine}}\) mask $M_i$ for strict inference constraints. The ego vehicle decodes the highly compressed payload (VQ indices and spatial headers) to reconstruct $M_i$ and the dense feature map $\hat{\mathbf{F}}$ for fusion (\cref{sec:coop_fuse}). The dashed arrow ($\bar{d}_{h,w}$) denotes the regularized offline supervision target (\cref{sec:train_obj}).}
  \label{fig:arch}
\end{figure*}

We propose a coverage-refinement communication design for intermediate-fusion cooperative perception. Each transmitting agent encodes its point cloud into a BEV feature map $\mathbf{F}\in\mathbb{R}^{C\times H\times W}$ and constructs a two-layer message: a dense coarse base layer over the full BEV map and sparse fine refinements selected under a strict byte budget. The receiver reconstructs a unified dense feature map $\hat{\mathbf{F}}$, allowing standard BEV fusion modules to operate without sparse-specific validity masks. When integrated into CoBEVT~\cite{xu_cobevt_2022}, we denote the resulting model as \textbf{CoBEVT-DR}. Unless stated otherwise, we describe the transmission of one non-ego agent and omit the agent index \(i\) for readability. In figures and multi-agent notation, \(\mathbf{F}_i\), \(\mathbf{B}_i\), \(\mathbf{M}^{\text{soft}}_i\), \(\mathbf{M}_i\), and \(\hat{\mathbf{F}}_i\) denote the BEV feature map, benefit map, soft training mask, deterministic inference mask, and reconstructed feature map of agent \(i\), respectively.

\subsection{Coverage-Refinement Feature Encoding}
\label{sec:met_feature}

We partition $\mathbf{F}$ into a grid of $G_h\times G_w$ non-overlapping cells of size $C_{\text{cell}}\times C_{\text{cell}}$. We use \((u,v)\) to denote dense BEV feature locations in the full \(H\times W\) map and \((h,w)\) to denote cell indices in the \(G_h\times G_w\) refinement grid. The set \(\Omega_{h,w}\) contains all dense BEV locations belonging to cell \((h,w)\). A cell is the atomic unit for fine-refinement selection, while VQ tokens are the atomic transmitted units. This keeps the fine-patch cost constant and avoids the coordinate overhead and fragmented context of selecting individual BEV locations.

Compression is performed with a shared Vector-Quantized Variational Autoencoder (VQ-VAE) codebook $\mathcal{C}\in\mathbb{R}^{K\times D}$. For each latent token $\mathbf{f}_{h,w,t}\in\mathbb{R}^{D}$, quantization selects the nearest codebook vector:
\begin{equation}
k_{h,w,t} = \mathop{\mathrm{arg\,min}}_{k \in \{1, \dots, K\}}
\|\mathbf{f}_{h,w,t} - \mathbf{e}_k\|_2^2,\qquad
\mathbf{e}_k\in\mathcal{C}.
\end{equation}
We use SimVQ~\cite{zhu_addressing_2024} for stable codebook utilization, but the coverage-refinement allocation is codec-independent: the codec maps transmitted representations to compact indices, while our policy determines where fine resolution is spent.

\begin{itemize}
    \item \textbf{Fine refinement.} A selected cell is transmitted at full spatial resolution, yielding
    $n_{\text{fine}}=(C_{\text{cell}}/s_{\text{enc}})^2$ tokens.
    \item \textbf{Coarse base layer.} The full BEV map is downsampled by $S_{\text{base}}$ before quantization and upsampled after decoding, providing dense low-rate coverage. For accounting on the same cell grid, each cell-equivalent region contributes
    $n_{\text{coarse}}\approx n_{\text{fine}}/S_{\text{base}}^2$ tokens.
\end{itemize}

\begin{figure}[h]
    \centering
    \includegraphics[width=\linewidth]{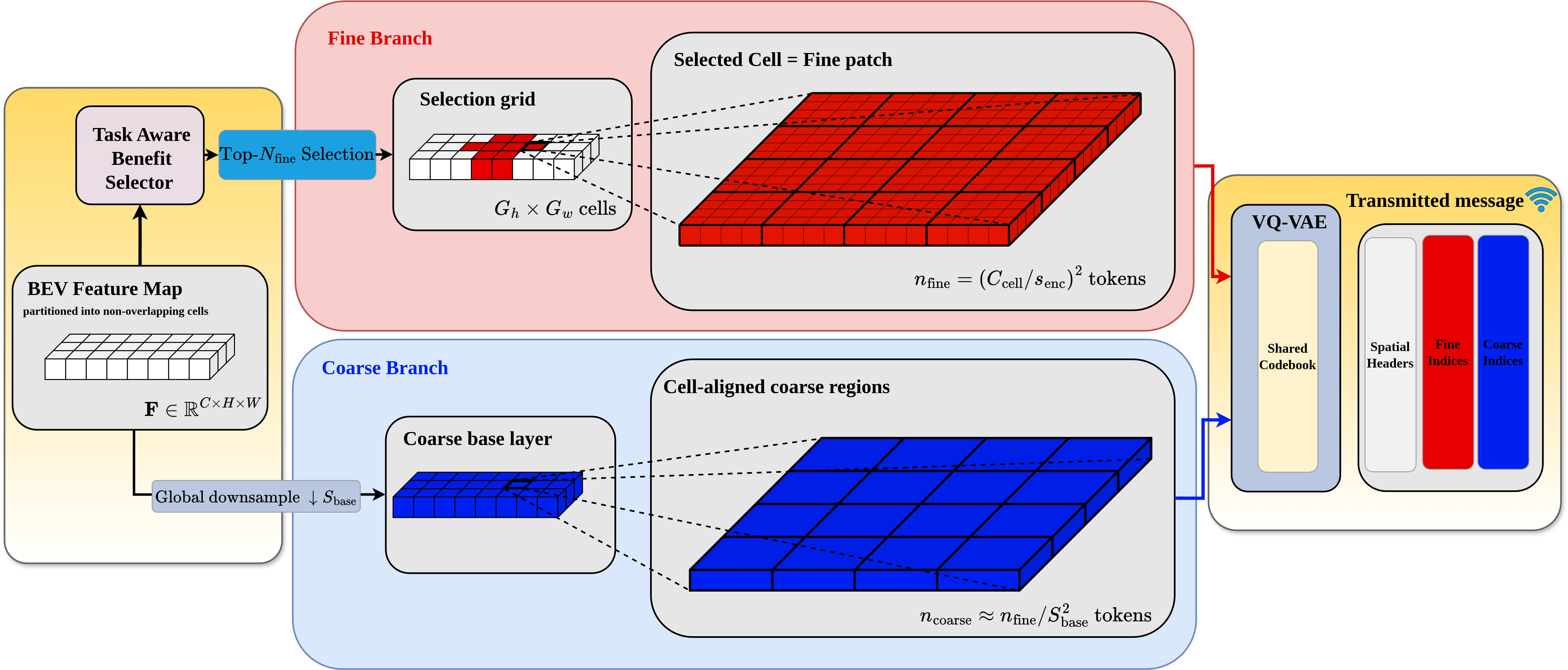}
\caption{\textbf{Spatial hierarchy and bitstream composition.}
The selector sends selected cells as fine patches, while a globally downsampled BEV map provides the dense coarse base layer. Both branches share the VQ-VAE codebook; the bitstream stores spatial headers, fine indices, and coarse indices.}
    \label{fig:unit_hierarchy}
\end{figure}

Both branches share the same VQ-VAE weights, so decoded coarse and fine features remain in a unified feature space. With $s_{\text{enc}}=1$, $C_{\text{cell}}=16$, and $S_{\text{base}}=4$, a selected fine cell contains $16\times16=256$ tokens, while its cell-equivalent coarse region contains $4\times4=16$ tokens. In \cref{sec:experiments}, we isolate the coverage-refinement design from quantization by comparing raw full, raw coarse, and raw coarse+fine transmission. \cref{fig:unit_hierarchy} illustrates the hierarchy between BEV locations, fine-selection cells, the globally downsampled coarse base layer, and VQ tokens.

\subsection{Task-Aware Benefit Prediction}
\label{sec:met_task_aware}

\textbf{Task Confidence Prior.}
We derive task confidence from channel-wise mean absolute BEV activation. For each feature location $(u,v)$:
\begin{equation}
m_{u,v}=\frac{1}{C}\sum_{c=1}^{C}\left|\mathbf{F}_{c,u,v}\right|, 
\Psi_{u,v}=\mathrm{clip}\left(\frac{m_{u,v}-p_{02}}{p_{98}-p_{02}+\epsilon},0,1\right),
\end{equation}
where $p_{02}$ and $p_{98}$ are per-vehicle 2nd/98th percentiles of $m$.
Each cell $(h,w)$ corresponds to a set of dense coordinates $\Omega_{h,w}$. We pool the dense prior to a cell-level prior:
\begin{equation}
\psi_{h,w} = \max_{(u,v) \in \Omega_{h,w}} \Psi_{u,v}.
\end{equation}

\textbf{Multi-Scale Benefit Prediction.}
All attention operations are performed at the full feature resolution. Our selector network $f_\theta$ processes the concatenated dense input $[\mathbf{F}, \Psi]$ using Pyramid Window Attention with parallel window sizes ($4\times4$, $8\times8$, and $16\times16$) in feature pixels. These representations are aggregated via Split Attention Fusion~\cite{zhang2022resnest} and projected to predict a dense distortion-reduction logit map $\delta_{u,v}$. We then pool to the cell grid as $\delta_{h,w}=\max_{(u,v)\in\Omega_{h,w}}\delta_{u,v}$, yielding per-cell scores under a shared target budget $B_{\text{target}}$. 
The final benefit score is
\begin{equation}
\beta_{h,w}=\max\left(\psi_{h,w}\cdot\left[w_{\text{task}}+(1-w_{\text{task}})\cdot\sigma(\delta_{h,w})\right], \beta_{\text{floor}}\right).
\end{equation}

where $\delta_{h,w}$ predicts a distortion-reduction proxy for upgrading cell $(h,w)$ from coarse to fine transmission, and $\sigma(\cdot)$ is the sigmoid function. The hyperparameter $w_{\text{task}}$ balances the task confidence prior against pure reconstruction fidelity, while $\beta_{\text{floor}}$ ensures a minimum selection probability for background regions. Crucially, this scalar score, $\beta_{h,w}$, allows us to globally rank and prioritize regions, enabling the system to adaptively allocate bandwidth within any target budget $B_{\text{target}}$. The score $\beta_{h,w}$ is used only for ranking cells under a byte budget; it does not require transmitting a dense score map to the receiver.
\subsection{Differentiable Selection}
\label{sec:differentiable_selection}
To enable end-to-end training of a discrete transmission policy, we use a Binary-Concrete relaxation~\cite{maddison2016concrete,jang2016categorical}. We collect the predicted benefits into a pragmatic benefit map $B_i=[\beta_{h,w}]\in\mathbb{R}^{G_h\times G_w}$. For each vehicle, the entries of $B_i$ are min-max normalized and median-centered to yield $\tilde{\beta}_{h,w}$. 
Let $B_{\text{coarse}}$ denote the fixed coarse-layer payload and $B_{\text{fine}}=B_{\text{target}}-B_{\text{coarse}}$ the remaining refinement budget. We then form selection logits $u_{h,w}=\tilde{\beta}_{h,w}-\lambda\cdot\rho_{h,w}$, where $\rho_{h,w}=b_{h,w}/B_{\text{fine}}$ is the normalized transmission cost.
 To allow differentiable sampling, we inject logistic noise $g_{h,w}\sim\mathrm{Logistic}(0,1)$ and compute a soft selection mask $M^{\text{soft}}_{h,w}=\sigma((u_{h,w}+g_{h,w})/\tau)$, where $\tau$ is linearly annealed during training to mitigate train-test distribution shift.

Here $M^{\text{soft}}\in\mathbb{R}^{G_h\times G_w}$ is applied to $\mathbf{F}_{\text{fine}},\mathbf{F}_{\text{coarse}}\in\mathbb{R}^{C\times H\times W}$ by tiling each mask entry over its corresponding spatial region $\Omega_{h,w}$ (and broadcasting across channels). The reconstructed feature map is a soft blend of both modes:
\begin{equation}
\hat{\mathbf{F}}=(1-M^{\text{soft}})\odot\mathbf{F}_{\text{coarse}}+M^{\text{soft}}\odot\mathbf{F}_{\text{fine}}.
\end{equation}

At inference, each cell has a predicted benefit $\beta_{h,w}$. As described in Supp. Sec.~D, the fine-patch payload is constant, i.e., $b_{h,w}=B_{\text{patch}}$ (e.g., 1{,}584 bits) for all candidate cells. Under this uniform-cost setting, our per-vehicle budgeted
selection formulation becomes a 0-1 knapsack problem with identical weights. 
This is mathematically equivalent to selecting the Top-\(N_{\text{fine}}\) cells ranked by \(\beta_{h,w}\), where \(N_{\text{fine}}=\left\lfloor(B_{\text{target}}-B_{\text{coarse}})/B_{\text{patch}}\right\rfloor\). Since all fine cells have identical cost, greedy Top-\(N_{\text{fine}}\) selection is optimal for the per-agent refinement budget. Moreover, decreasing the budget simply truncates the same ranked list, producing nested selections across budgets. This property enables zero-retraining budget adaptation at inference. To maximize bandwidth efficiency, the final deterministic mask \(\mathbf{M}_i\) is not transmitted as a dense binary grid; rather, the coordinates of the selected Top-\(N_{\text{fine}}\) cells are extracted and transmitted as per-patch spatial headers, allowing the receiver to reconstruct \(\mathbf{M}_i\).
\subsection{Budget-Constrained Training}
\label{sec:met_dual_asc}
During training, the dense coarse base layer has a fixed payload \(B_{\text{coarse}}\), so the remaining refinement budget is \(B_{\text{fine}}=B_{\text{target}}-B_{\text{coarse}}\). The soft mask induces an expected fine-refinement payload \(\hat{B}_{\text{fine}}=\sum_{h,w}M^{\text{soft}}_{h,w}b_{h,w}\). We penalize budget violations by comparing \(\hat{B}_{\text{fine}}\) against \(B_{\text{fine}}\): \begin{equation} \mathrm{pen}(\hat{B}_{\text{fine}}) = 2\cdot \mathrm{ReLU}\left(\frac{\hat{B}_{\text{fine}}}{B_{\text{fine}}}-1\right) + \mathrm{ReLU}\left(0.8-\frac{\hat{B}_{\text{fine}}}{B_{\text{fine}}}\right). \end{equation}
The first term strongly discourages over-budget messages, while the second avoids trivial under-use of the channel. The penalty weight is controlled by a dual variable $\lambda$, updated by projected dual ascent during training. In inference, no soft penalty is needed because deterministic Top-\(N_{\text{fine}}\) selection exactly satisfies the target byte budget.

\subsection{Training Objectives}
\label{sec:train_obj}
The system is trained end-to-end from scratch, minimizing a joint objective that comprises detection accuracy, codebook commitment, selector supervision, and the dynamic budget penalty:
\begin{equation}
\mathcal{L}=\mathcal{L}_{\text{det}}+\alpha_{\text{vq}}\mathcal{L}_{\text{VQ}}+\alpha_{\text{rd}}\mathcal{L}_{\text{sup}}+\mathcal{L}_{\text{budget}},
\end{equation}

\textbf{Detection ($\mathcal{L}_{\text{det}}$) \& Commitment Losses ($\mathcal{L}_{\text{VQ}}$)}. We apply standard detection losses (Focal Loss for classification heatmaps and Smooth-L1 for bounding box regression parameters) to the final fused feature map. To ensure continuous embeddings align stably with the discrete codebook, we minimize the standard VQ commitment loss~\cite{oord_neural_2018} alongside our joint objective.

\noindent\textbf{Benefit supervision.}
To train the selector, we construct an offline target measuring the relative gain of fine over coarse reconstruction:
\begin{equation}
\Delta d_{h,w}=
\log\left(\|\mathbf{F}_{h,w}-\mathbf{F}_{\text{coarse},h,w}\|_2^2\right)
-
\log\left(\|\mathbf{F}_{h,w}-\mathbf{F}_{\text{fine},h,w}\|_2^2\right).
\end{equation}
We normalize this value to $\bar{d}_{h,w}\in[0,1]$ and gate it by the task prior $\psi_{h,w}$:
\begin{equation}
\beta^*_{h,w} =
\max\!\left(
\psi_{h,w}\left[w_{\text{task}}+(1-w_{\text{task}})\bar{d}_{h,w}\right],
\beta_{\text{floor}}
\right).
\end{equation}
This suppresses high log-ratio values in empty or noisy regions and encourages refinement where reconstruction gain is also task-relevant. The selector is trained with Smooth-L1 losses on both $\beta_{h,w}$ and the auxiliary distortion term $\sigma(\delta_{h,w})$, with stop-gradient applied to the targets.
\subsection{Reconstruction and Cooperative Fusion}
\label{sec:coop_fuse}
At the receiver, coarse and selected fine indices are decoded with the shared VQ codebook. The transmitted patch headers reconstruct the refinement mask, which replaces the corresponding coarse regions with fine-resolution features to obtain a dense map $\hat{\mathbf{F}}$. Neighbor features are then warped into the ego frame and fused with a standard BEV fusion module. We use SwapFusion~\cite{xu_cobevt_2022} by default and additionally evaluate MaxFusion and Where2comm-style fusion in \cref{sec:generalization_bandwidth}. Since reconstruction is dense, no sparse-specific validity masks or fusion modifications are required.

\section{Evaluation}
\label{sec:experiments}

\subsection{Implementation Details}
\label{sec:impl_details}
We evaluate on the real-world DAIR-V2X~\cite{yu_dair-v2x_2022} and simulated OPV2V~\cite{xu_opv2v_2022} benchmarks, which represent complementary V2I and V2V cooperative-perception settings~\cite{yazgan_datasets_2024}. Following standard cooperative-perception settings~\cite{mmcooper,xu_cosdh_2025,infocom}, PointPillars voxelization uses a $(0.4\,\mathrm{m},0.4\,\mathrm{m},4.0\,\mathrm{m})$ grid, and the encoder outputs the intermediate BEV feature map $\mathbf{F}$ after a $4\times$ spatial downsampling stride. DAIR-V2X annotations are extended following common practice~\cite{coalign}. Unless stated otherwise, controlled variants share the same backbone, detection head, fusion module, training protocol, and evaluation code, differing only in the transmitted representation.

Payload is reported as the \emph{decodable per-agent message size}, i.e., the bit-packed representation including all VQ indices and side information required for reconstruction; dense masks and benefit maps are not transmitted. We use 1 KB = 1,000 bytes. CoBEVT-DR reports realized payload under a strict per-message cap, whereas threshold-based baselines use the dataset-average operating point closest to the target budget. Our SimVQ codebook uses $K=64$ entries (6 bits/token) with latent dimension $d=64$. The BEV map is partitioned into $16\times16$ cells; each selected fine patch contains 256 tokens (1,536 bits), while the $S_{\text{base}}=4$ coarse layer contributes 16 tokens (96 bits) per cell-equivalent region. Each fine patch also carries a 48-bit spatial header including coordinates, framing, and CRC-16. Full accounting is provided in Supp. Sec.~D.

For reproduced baselines, we use official implementations and released fusion modules when available. If a method is incompatible with our controlled OpenCOOD/PointPillars+SwapFusion setup, we evaluate its communication module within the same detector and fusion pipeline to isolate communication efficiency. Megabyte-scale baselines are omitted from OPV2V kilobyte-regime comparisons; dense CoBEVT is reported as the dense uncompressed reference. Models are trained for 30 epochs on one RTX 4090, with batch size 4 for DAIR-V2X and 2 for OPV2V. Hyperparameters and additional codec studies are given in Supp. Secs.~A and~B.

\subsection{Controlled Diagnostics}
\label{sec:controlled_diagnostics}

Before comparing against external baselines, we isolate the source of the communication gain through two controlled diagnostics: a frozen-checkpoint selector isolation and a no-quantization feature diagnostic.
\begin{table}[h!]
\centering
\scriptsize
\caption{\textbf{Controlled diagnostics on DAIR-V2X.}
(A) Frozen-checkpoint selector isolation under the same 2~KB cap. 
(B) No-VQ diagnostic separating coverage-refinement allocation from quantization. Payload is per non-ego agent; C/F denotes coarse/fine.}
\label{tab:controlled_diagnostics}
\setlength{\tabcolsep}{4pt}
\begin{tabular}{l l c c c}
\toprule
\textbf{Test} & \textbf{Variant} & \textbf{AP@0.5}$\uparrow$ & \textbf{AP@0.7}$\uparrow$ & \textbf{KB}$\downarrow$ \\
\midrule
\multirow{3}{*}{(A) Frozen ranking}
& Random & 0.680 & 0.573 & 1.878 \\
& Prior only ($\psi$) & 0.713 & 0.589 & 1.878 \\
& Learned benefit ($\beta$) & \textbf{0.736} & \textbf{0.604} & 1.878 \\
\midrule
\multirow{4}{*}{(B) No VQ}
& Raw full & 0.720 & 0.580 & 6291.46 \\
& Raw C & 0.651 & 0.546 & 393.22 \\
& Raw C+F (25\% fine cells) & 0.715 & 0.583 & 1966.12 \\
& DR+SimVQ & \textbf{0.736} & \textbf{0.604} & \textbf{1.878} \\
\bottomrule
\end{tabular}

\vspace{2pt}
\begin{minipage}{0.98\linewidth}
\raggedright
\emph{Note:} In the no-VQ diagnostic, Raw C+F uses the dense raw coarse base plus the same selected fine-cell ratio as the 2~KB DR setting, but transmits raw features without quantization.
\end{minipage}
\end{table}
\paragraph{Frozen-checkpoint selector isolation.}
\cref{tab:controlled_diagnostics}(A) isolates the ranking policy from codec training. All variants use the same frozen encoder, SimVQ codebook, decoder, detection head, coarse layer, and 2~KB payload cap; only the inference-time ranking score is changed. Thus, the quantizer cannot be specialized differently for random, prior-only, or learned selection. The learned benefit score improves over both random selection and the task-prior-only ranking at the same 1.878~KB payload, indicating that the selector improves the allocation of fixed codec capacity rather than relying only on codec-selector co-adaptation.

\paragraph{No-quantization diagnostic.}
\cref{tab:controlled_diagnostics}(B) removes SimVQ from the diagnostic variants. Raw coarse transmission provides dense coverage but loses fine detail, while raw coarse+fine recovers near full-raw accuracy at substantially lower payload. Thus, the dual-resolution layout contributes independently of quantization. SimVQ then reduces this MB-scale coverage-refinement representation to the final kilobyte regime.
\subsection{Comparison with Communication-Efficient Baselines}
\label{sec:baseline_comparison}

\cref{tab:baseline_compare} compares detection accuracy and communication cost under ideal pose. Since cooperative-perception results are sensitive to dataset split, annotation extension, spatial range, backbone, fusion module, training schedule, and payload accounting, we separate protocol-controlled comparisons from external baselines. The controlled CoBEVT codec variants use the same OpenCOOD/CoBEVT pipeline, detector, fusion module, training protocol, and evaluation code, isolating the effect of the transmitted representation. External baselines are reported using released implementations when compatible; otherwise, controlled reproductions are explicitly marked. Payload is measured as average decodable KB per non-ego agent per frame, including all side information required for receiver reconstruction. For sparse baselines, masks or selected regions are encoded in the most compact decodable form supported by the implementation, using coordinates or bit-packed indices when available rather than assuming dense floating-point score maps. For threshold-based masking methods, we sweep the masking threshold and report the operating point whose mean payload is closest to our target budget. In contrast, CoBEVT-DR enforces a strict byte cap through ranked truncation.
\begin{table}[h!]
\centering
\scriptsize
\caption{\textbf{Accuracy-payload comparison under ideal pose.} Payload (KB) is the average decodable message size per non-ego agent. Threshold-based methods report the closest mean payload from a sweep. OPV2V is denser; MB-scale baselines are omitted from the kilobyte-regime comparison and shown as N/A.}
\label{tab:baseline_compare}
\begin{threeparttable}
\resizebox{0.98\columnwidth}{!}{%
\begin{tabular}{l c c r c c r}
\toprule
\multirow{2}{*}{\textbf{Method}} & \multicolumn{3}{c}{\textbf{DAIR-V2X}~\cite{yu_dair-v2x_2022}} & \multicolumn{3}{c}{\textbf{OPV2V}~\cite{xu_opv2v_2022}} \\
\cmidrule(lr){2-4} \cmidrule(lr){5-7}
 & AP@0.5$\uparrow$ & AP@0.7$\uparrow$ & KB$\downarrow$ & AP@0.5$\uparrow$ & AP@0.7$\uparrow$ & KB$\downarrow$ \\
\midrule
\multicolumn{7}{l}{\textit{Dense \& Sparse Baselines (High Bandwidth)}} \\
CoBEVT (Original) & 0.72 & 0.58 & 6291.46 & 0.95 & 0.88 & 8650 \\
Where2comm~\cite{hu_where2comm_2022} & 0.72 & 0.57 & 2415.22 & \multicolumn{3}{c}{N/A (MB-scale)} \\
ERMVP~\cite{ermvp} & 0.70 & 0.57 & 1230.00 & \multicolumn{3}{c}{N/A (MB-scale)} \\
EffiComm~\cite{Yazgan_2025_ITSC} & 0.72 & 0.57 & 785.00 & \multicolumn{3}{c}{N/A (MB-scale)} \\
\midrule
\multicolumn{7}{l}{\textit{Efficient \& Quantized Approaches (Low Bandwidth)}} \\
CodeFilling~\cite{hu_communication-efficient_2024} & 0.74 & 0.59 & $\sim$27.00 & 0.92 & 0.85 & $\sim$12.00 \\
CoSDH\textsuperscript{*}~\cite{xu_cosdh_2025} & 0.67 & 0.54 & $\sim$4.63 & 0.85 & 0.73 & $\sim$3.3 \\
mmCooper~\cite{mmcooper} & 0.70 & 0.57 & $\sim$3.80 & 0.95 & 0.88 & $\sim$9.40 \\
InfoCom\textsuperscript{+} (SwapFusion)~\cite{infocom} & 0.66 & 0.55 & $\sim$2.32 & 0.95 & 0.86 & $\sim$2.92 \\
\midrule
\multicolumn{7}{l}{\textit{Controlled CoBEVT Codec Variants}} \\
CoBEVT-RVQ$^\dagger$ (3-stage, $K=64$) & 0.70 & 0.54 & 13.82 & 0.94 & 0.86 & 19.2 \\
CoBEVT-SimVQ ($K=64$) & 0.71 & 0.52 & 4.61 & 0.94 & 0.85 & 6.41 \\
\textbf{CoBEVT-DR} & 0.74 & 0.60 & 1.87 & 0.95 & 0.87 & 1.98 \\
\bottomrule
\end{tabular}%
}
\begin{minipage}{\columnwidth}
\scriptsize
\raggedright
\setlength{\parindent}{0pt}
\textit{Notes.}
\textsuperscript{+} \textbf{InfoCom} is integrated into our PointPillars+SwapFusion pipeline because the released fusion stack is incompatible with our controlled OpenCOOD setup. We report decodable payload, including latent features, quantized masks, and bit-packed indices; see Supp.\ Sec.~D.

\textsuperscript{*} \textbf{CoSDH} is request-response by design. We report a budget-adjusted operating point obtained by sweeping the communication threshold. Our reproduction of the original high-payload operating point reaches 0.773 AP@0.5 and 0.598 AP@0.7 on DAIR-V2X, but requires 174.53~KB per non-ego agent; this full operating point is included in \cref{fig:robustness_pose}. To avoid penalizing CoSDH for its query stage, all CoSDH payloads report only the response payload from non-ego agent to ego, following~\cite{xu_cosdh_2025}; see Supp.\ Sec.~D.

$^\dagger$ \textbf{CoBEVT-RVQ} is our reproduction of an RVQ-based baseline in the same CoBEVT pipeline, following~\cite{revqom2025}, for which no official implementation is available.
\end{minipage}
\end{threeparttable}
\end{table}
Within the controlled codec comparison, increasing quantization rate
alone does not recover the same trade-off. Uniform SimVQ reaches
0.52 AP@0.7 at 4.61 KB and three-stage RVQ reaches 0.54 at
13.82 KB, whereas CoBEVT-DR achieves 0.60 AP@0.7 at only
1.87 KB. This shows that task-aware spatial allocation is substantially
more effective than uniformly increasing codec rate in the kilobyte regime. Additional codec studies in Supp. Sec.~B support this design choice: global downsampling strongly degrades high-IoU precision, Finite Scalar Quantization (FSQ) and AutoEncoder (AE) require substantially larger payloads, and RVQ improves accuracy mainly by increasing the number of transmitted stages. We do not interpret the small gain over dense CoBEVT on DAIR-V2X as evidence that lossy compression is generally superior to dense transmission; rather, uniform SimVQ and RVQ remain below the dense reference, and the gain appears only when bytes are allocated through the task-aware coverage-refinement policy. Compared with low-bandwidth external baselines under the stated protocols, CoBEVT-DR provides a strong accuracy-payload trade-off on DAIR-V2X and remains close to dense CoBEVT-level accuracy on OPV2V while using only kilobyte-scale messages. Additional qualitative comparisons are provided in Supp. Sec.~H.
\subsection{Generalization and Dynamic Bandwidth}
\label{sec:generalization_bandwidth}
We evaluate whether the proposed coverage-refinement principle transfers across
fusion modules, LiDAR backbones, and downstream BEV tasks, and whether a single
trained model adapts to changing byte budgets without retraining.
Dynamic-object BEV segmentation results are provided in Supp.\ Sec.~I, while
additional V2X-Real~\cite{xiang2024v2x} and V2XVerse~\cite{codriving} experiments on multi-class detection are reported in Supp.\ Sec.~J and K.
\paragraph{Fusion-module generalization.}
To test whether the proposed message construction and allocation design depends on CoBEVT's attention fusion, we integrate the same coverage-refinement communication design into non-attention MaxFusion and Where2comm~\cite{hu_where2comm_2022} multi-scale fusion. For Where2comm, we disable its original sparse communication mechanism during training and replace the transmitted message with our dense coarse plus sparse refinement representation. As shown in~\cref{tab:fusion_generalization}, DR improves AP@0.7 under both fusion settings while reducing payload to the kilobyte regime. This indicates that the proposed communication design is not tied to a specific fusion operator.

\begin{table}[h!]
\centering
\scriptsize
\caption{\textbf{Fusion generalization on DAIR-V2X.} DR improves high-IoU accuracy with MaxFusion and Where2comm-style fusion while reducing payload to the kilobyte regime. Payload is per non-ego agent.}
\label{tab:fusion_generalization}
\resizebox{\linewidth}{!}{%
\begin{tabular}{l l c c c}
\toprule
\textbf{Fusion} & \textbf{Communication} & \textbf{AP@0.5}$\uparrow$ & \textbf{AP@0.7}$\uparrow$ & \textbf{KB}$\downarrow$ \\
\midrule
MaxFusion & SimVQ & 0.686 & 0.457 & 4.61\\
MaxFusion & DR & 0.682 & \textbf{0.525} & \textbf{1.878} \\
\midrule
Where2comm & SimVQ & 0.700 & 0.532 & 4.61 \\
Where2comm & DR & \textbf{0.721} & \textbf{0.575} & \textbf{1.878} \\
\bottomrule
\end{tabular}%
}
\end{table}

\begin{figure}[h!] 
    \centering
    
    \begin{subfigure}[b]{0.48\linewidth}
        \centering
        \includegraphics[width=\linewidth]{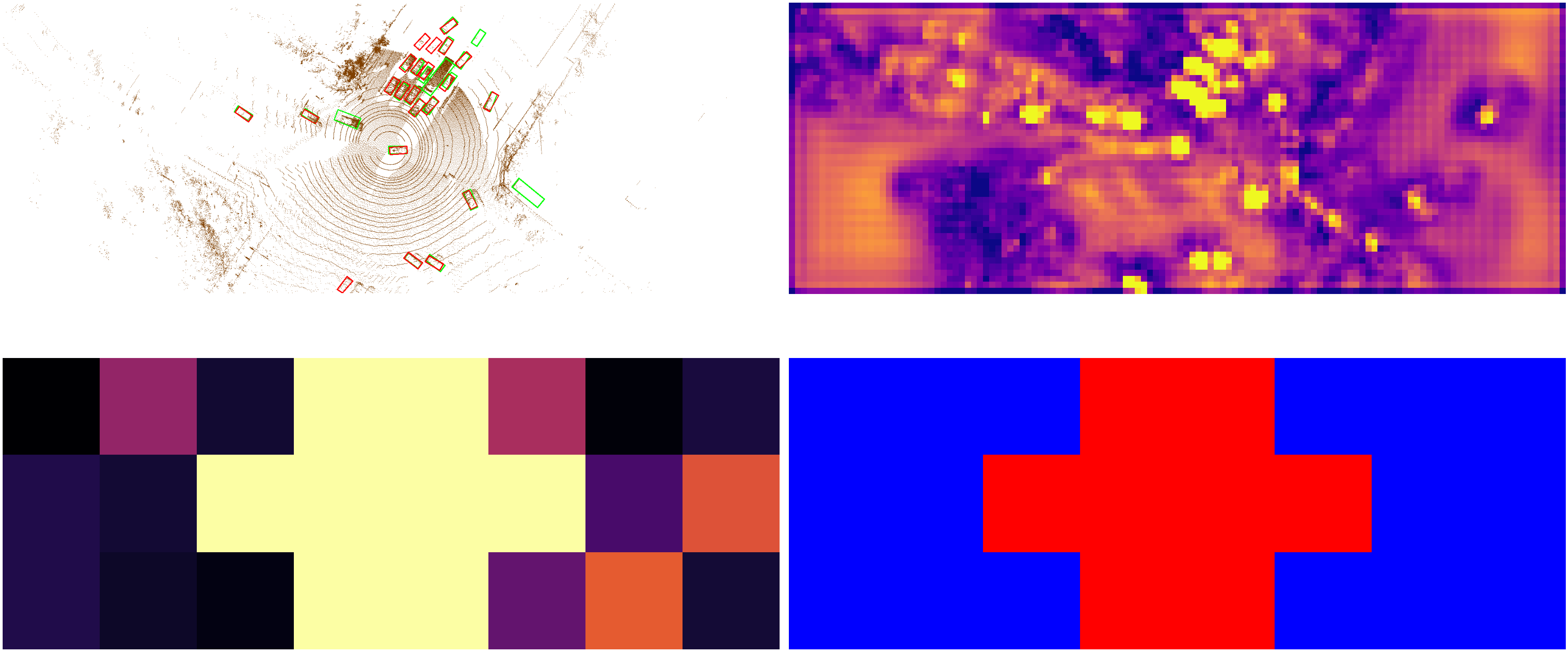}
        \caption{\textbf{Task-Aware Pipeline}: Scene context mapped to Benefit Map and selection mask. \textbf{Red}: Fine patches; \textbf{Blue}: Coarse floor.}
        \label{fig:selection_policy}
    \end{subfigure}
    \hfill
    \begin{subfigure}[b]{0.48\linewidth}
        \centering
        \includegraphics[width=\linewidth]{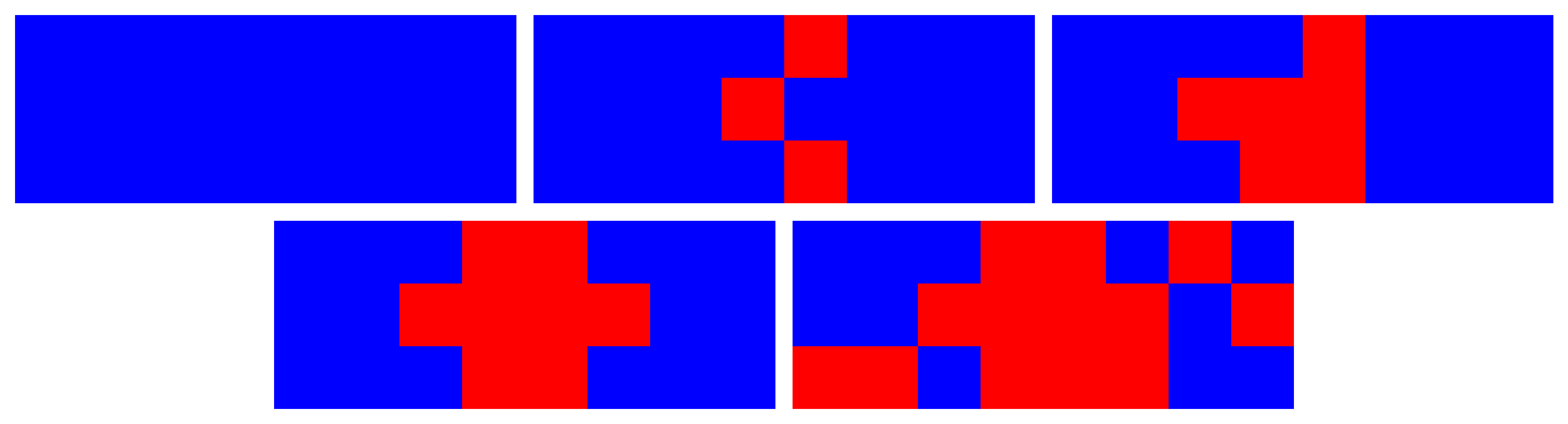}
        \caption{\textbf{Monotonic Scalability}: Policy expands strictly from object cores to context as budget increases.}
        \label{fig:budget_sweep}
    \end{subfigure}

    \caption{\textbf{Integrated Scalability Analysis.} Visual logic \subref{fig:selection_policy} and mask expansion \subref{fig:budget_sweep} illustrate the learned refinement policy used for fixed and dynamic budget adaptation.}
    \label{fig:combined_scalability_analysis}
\end{figure}

\paragraph{Backbone generalization.}
We replace only the LiDAR backbone with SECOND~\cite{second}, while keeping the communication head, Task-Aware Benefit Selector, and SwapFusion module unchanged. The communicated tensor maintains the same dense BEV interface, with \(C=256\) channels and a spatial size of \(48\times128\); therefore, the same cell decomposition, tokenization, and payload accounting apply. As shown in~\cref{tab:second_backbone_generalization}, DR improves SECOND+SimVQ from 0.681 to 0.707 AP@0.5 and from 0.404 to 0.518 AP@0.7 at a lower payload. This suggests that the gain comes from the coverage-refinement communication design rather than a PointPillars-specific interaction.
\begin{table}[h!]
    \centering
    \caption{\textbf{Backbone generalization with SECOND on DAIR-V2X.}
DR improves over uniform SimVQ with the SECOND backbone at lower payload. Payload is per non-ego agent.}
    \label{tab:second_backbone_generalization}
    \scriptsize
    \setlength{\tabcolsep}{6pt}
    \begin{tabular}{l c c c}
        \toprule
        Method & AP@0.5$\uparrow$ & AP@0.7$\uparrow$ & KB$\downarrow$ \\
        \midrule
        SECOND & 0.712 & 0.499 & 6291.46 \\
        SECOND + SimVQ & 0.681 & 0.404 & 4.61 \\
        SECOND + Ours & 0.707 & 0.518 & 1.878 \\
        \bottomrule
    \end{tabular}
\end{table}
\paragraph{Dynamic budget adaptation.}
Finally, we test whether one trained model can operate under changing byte budgets. As established in~\cref{sec:differentiable_selection}, constant fine-patch cost makes the selected patch set a nested ranked prefix. \cref{fig:combined_scalability_analysis} visualizes this behavior: the selector first allocates fine patches to high-value object cores and then expands toward surrounding context as the budget increases. Quantitatively, \cref{tab:budget_adaptation}(A) evaluates one trained model across fixed hard budget caps on DAIR-V2X. The coarse layer alone provides a dense fallback, and performance saturates around the 2.0~KB operating point. \cref{tab:budget_adaptation}(B) further evaluates per-frame, per-agent budget variation on OPV2V, where each agent receives a budget from $\{0,1,2\}$~KB and transmits the largest decodable prefix that fits the assigned cap. Under this dynamic schedule, the model maintains 0.920 AP@0.5 at only 1.213~KB average payload, demonstrating zero-retraining adaptation to abrupt bandwidth changes.
Multi-agent scaling with up to five non-ego collaborators is reported in Supp. Sec. E; under a 2.0 KB/agent cap, total communication remains only 10 KB at N=5, compared with $\sim31.4$ MB for dense exchange.

\begin{table}[h!]
    \centering
    \caption{\textbf{Budget adaptation.} 
    (A) Fixed-budget scalability on DAIR-V2X using a single trained model. 
    (B) Dynamic per-frame budget scheduling on OPV2V, where each agent receives a budget from \(\{0,1,2\}\) KB and 0 KB means no transmission.}
    \label{tab:budget_adaptation}
    \scriptsize
    \setlength{\tabcolsep}{5pt}

    \textbf{(A) Fixed budget scalability on DAIR-V2X}
    \vspace{1mm}

    \begin{tabular}{l c c c}
        \toprule
        Budget & AP@0.5$\uparrow$ & AP@0.7$\uparrow$ & R@0.5$\uparrow$ \\
        \midrule
        \textit{Coarse} & 0.645 & 0.554 & 0.670 \\
        1.0 KB & 0.706 & 0.588 & 0.743 \\
        1.5 KB & 0.730 & 0.601 & 0.773 \\
        \textbf{2.0 KB} & \textbf{0.736} & \textbf{0.604} & \textbf{0.779} \\
        3.0 KB & 0.738 & 0.605 & 0.782 \\
        \bottomrule
    \end{tabular}

    \vspace{2mm}

    \textbf{(B) Dynamic budget scheduling on OPV2V}
    \vspace{1mm}

    \begin{tabular}{l c c c c}
        \toprule
        Budget & AP@0.5$\uparrow$ & AP@0.7$\uparrow$ & R@0.5$\uparrow$ & KB$\downarrow$ \\
        \midrule
        1 KB & 0.934 & 0.853 & 0.943 & 0.99 \\
        2 KB & 0.955 & 0.874 & 0.966 & 1.98 \\
        0/1/2 KB & 0.920 & 0.830 & 0.932 & 1.213 \\
        \bottomrule
    \end{tabular}
\end{table}
\subsection{Robustness to Real-World Challenges}
\label{sec:robustness}

\begin{figure}[h!]
    \centering
    \begin{subfigure}{0.49\columnwidth}
        \includegraphics[width=\linewidth]{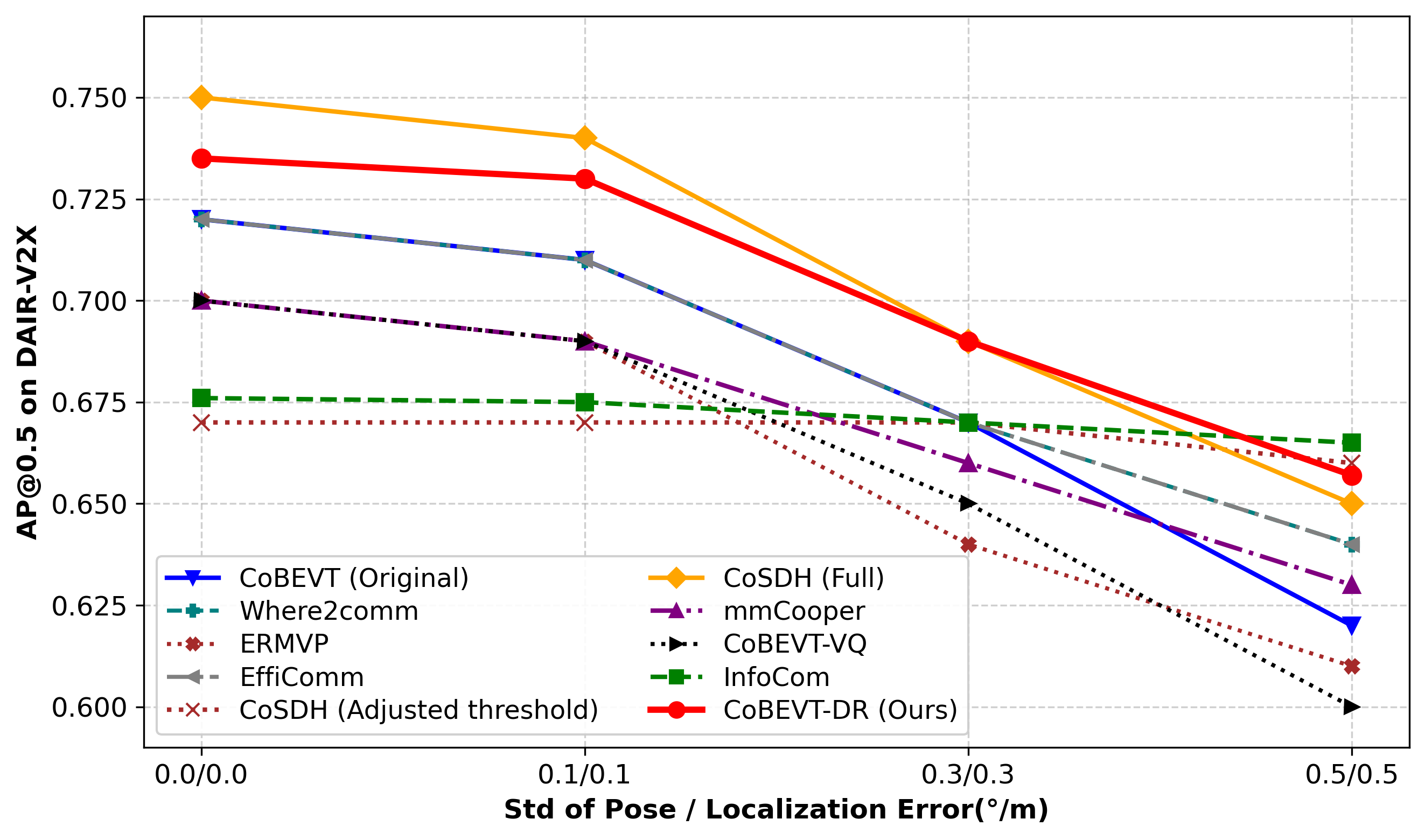} 
        \caption{Pose and Localization Error Robustness on DAIR-V2X Dataset} 
        \label{fig:robustness_pose}
    \end{subfigure}
    \hfill 
    \begin{subfigure}{0.49\columnwidth}
        \includegraphics[width=\linewidth]{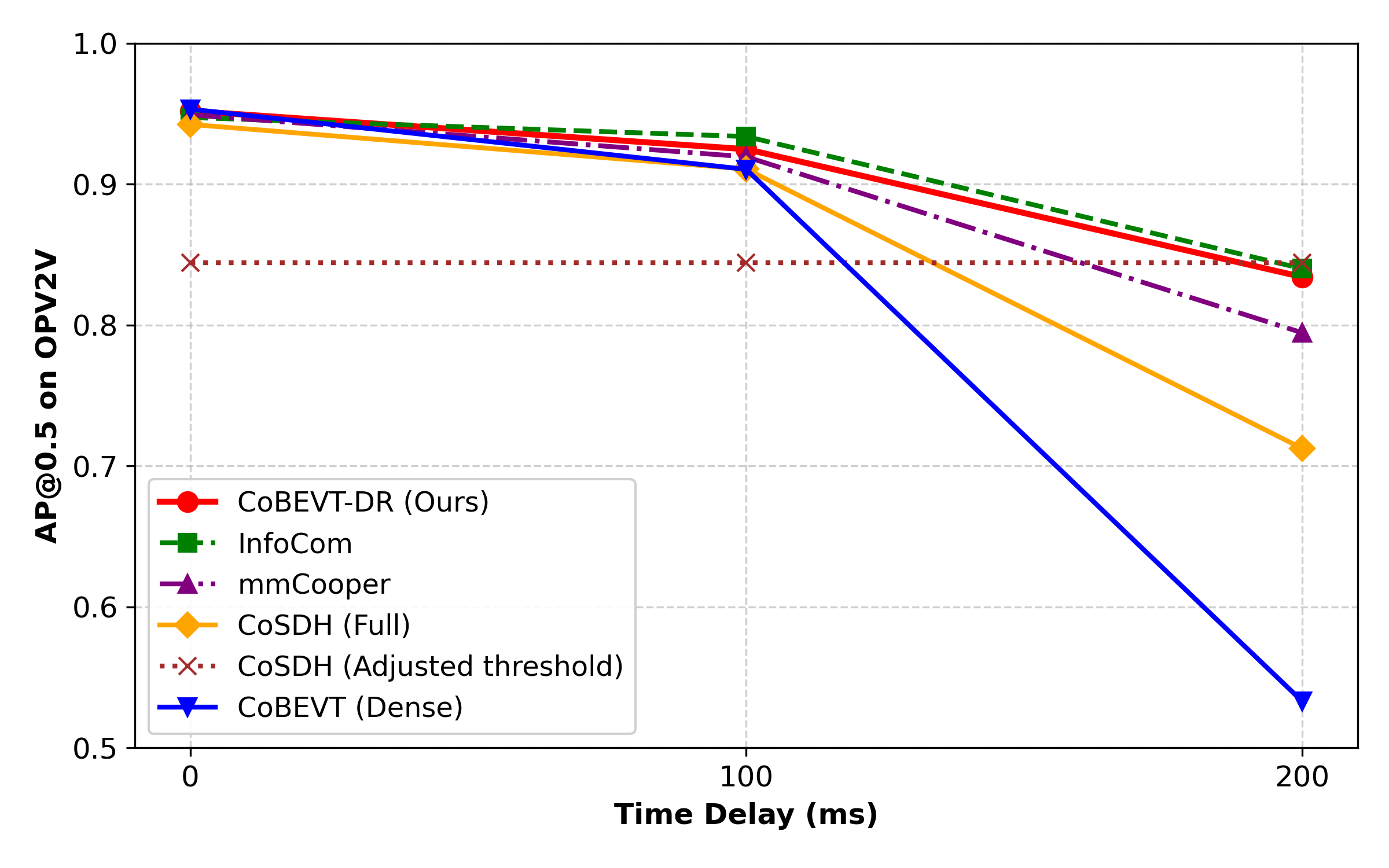} 
        \caption{Time Delay Robustness on OPV2V} 
        \label{fig:robustness_delay}
    \end{subfigure}
    \caption{\textbf{Robustness Analysis.} (a) \textbf{Pose Error}: High-bandwidth baselines (\textit{Full}) degrade rapidly under noise, while threshold-constrained versions (\textit{Adjusted}) suffer from low performance ceilings. (b) \textbf{Time Delays}: CoBEVT-DR and InfoCom show comparable stability, while CoBEVT-DR maintains a stronger clean-pose operating point.}
    \label{fig:real_world_robustness}
\end{figure}

Following the noise settings in~\cite{xu2022v2xvit,mmcooper}, composite localization and heading noise are sampled from Gaussian distributions $\mathcal{N}(0,\sigma_t)$ and $\mathcal{N}(0,\sigma_\theta)$, with $\sigma_t \in \{0.0,0.1,0.3,0.5\}$~m and $\sigma_\theta \in \{0.0,0.1,0.3,0.5\}^{\circ}$. We also introduce sender-side time delays to simulate asynchronous communication. \cref{fig:real_world_robustness} summarizes the results.
\paragraph{Localization and pose error.}
Dense fusion and uniform quantization achieve strong peak accuracy but are sensitive to spatial misalignment, while highly sparse baselines are more stable but have lower performance ceilings. CoBEVT-DR provides a middle ground: the coarse base layer supplies a continuous spatial anchor, while fine patches preserve high-utility details. Under severe noise ($0.5$~m/$0.5^{\circ}$), CoBEVT-DR remains around 0.65 AP@0.5 while retaining a stronger clean-pose operating point.

\paragraph{Asynchronous time delays.}
We evaluate temporal misalignment on OPV2V with delays up to 200~ms. Dense fusion degrades strongly as stale features are aggregated, whereas CoBEVT-DR maintains 0.834 AP@0.5 at 200~ms. Its stability is comparable to the aggressively sparse InfoCom baseline, but with a higher clean-pose accuracy. This indicates that the compressed coarse representation provides a useful regularized fallback under moderate asynchrony, although our method does not explicitly perform temporal compensation.
\paragraph{Real-time feasibility.}
Our one-shot broadcast design avoids the request-response handshakes required by interactive protocols such as CoSDH~\cite{xu_cosdh_2025}; detailed latency modeling is provided in Supp. Sec.~F.

\subsection{Ablation Studies}
\label{sec:ablation}

We validate the main architectural choices under strict bandwidth budgets on DAIR-V2X. \cref{tab:ablation_struct} studies the coarse-layer stride and cell size, \cref{tab:ablation_supervision} evaluates task-aware supervision, and \cref{tab:no_coarse} verifies the need for the dense coarse floor.

\paragraph{Structural ablation.}
We ablate the coarse-layer stride $S_{\text{base}}$ and selection granularity $C_{\text{cell}}$ in~\cref{tab:ablation_struct}. Setting $S_{\text{base}}=2$ consumes $\sim$1.16~KB of the budget for the base layer and leaves too little capacity for refinement, while $S_{\text{base}}=8$ is too coarse to provide a meaningful semantic floor. Our default $S_{\text{base}}=4$ provides the best trade-off. For granularity, coarser selection ($C_{\text{cell}}=32$) wastes bandwidth on background regions, while fine selection ($C_{\text{cell}}=8$) increases metadata overhead. We therefore use $C_{\text{cell}}=16$.
\begin{table}[h!]
\centering
\caption{\textbf{Structural ablation at 2.0 KB.}
Effect of base-layer stride $S_{\text{base}}$ and cell granularity $C_{\text{cell}}$.}
\label{tab:ablation_struct}
\scriptsize
\setlength{\tabcolsep}{3pt}
\begin{tabular}{l l c c c}
\toprule
\textbf{Param.} & \textbf{Set} & \textbf{Base}$\downarrow$ & \textbf{AP@0.5}$\uparrow$ & \textbf{AP@0.7}$\uparrow$ \\
\midrule
\multirow{3}{*}{$S_{\text{base}}$}
& $8$ & $\sim$0.08 & 0.726 & 0.594 \\
& $2$ & $\sim$1.16 & 0.730 & 0.593 \\
& \textbf{$4$} & \textbf{$\sim$0.29} & \textbf{0.736} & \textbf{0.604} \\
\midrule
\multirow{3}{*}{$C_{\text{cell}}$}
& $8$  & -- & 0.735 & 0.600 \\
& $32$ & -- & 0.718 & 0.589 \\
& \textbf{$16$} & -- & \textbf{0.736} & \textbf{0.604} \\
\bottomrule
\end{tabular}
\end{table}
\paragraph{Selector supervision.}
We compare task-aware selector supervision against a task-agnostic variant trained only on the internal reconstruction-improvement signal ($\Delta$MSE). This ablation tests whether selecting cells by reconstruction gain alone is sufficient, or whether refinement must be aligned with downstream detection utility. As shown in~\cref{tab:ablation_supervision}, task-aware supervision improves AP@0.5 by +3.3 points and AP@0.7 by +4.9 points at 2.0~KB. Selector architecture and utility diagnostics are provided in Supp. Secs. C and G. The adopted Pyramid Attention selector uses 1.27 M parameters and 28.84 GFLOPs, and its predicted benefit achieves a ROC-AUC of 0.785 for identifying high-utility patches according to counterfactual marginal detection loss.
\begin{table}[h!]
\centering
\caption{\textbf{Selector supervision ablation.}
Task-aware supervision $\Psi$ outperforms reconstruction-only supervision $\Delta$MSE.}
\label{tab:ablation_supervision}
\scriptsize
\setlength{\tabcolsep}{4pt}
\begin{tabular}{l cc cc}
\toprule
\multirow{2}{*}{\textbf{Target}} 
& \multicolumn{2}{c}{\textbf{1.0 KB}} 
& \multicolumn{2}{c}{\textbf{2.0 KB}} \\
\cmidrule(lr){2-3}\cmidrule(lr){4-5}
& \textbf{AP@0.5}$\uparrow$ & \textbf{AP@0.7}$\uparrow$
& \textbf{AP@0.5}$\uparrow$ & \textbf{AP@0.7}$\uparrow$ \\
\midrule
$\Psi$      & \textbf{0.706} & \textbf{0.588} & \textbf{0.736} & \textbf{0.604} \\
$\Delta$MSE & 0.676 & 0.535 & 0.703 & 0.555 \\
\bottomrule
\end{tabular}
\end{table}

\paragraph{No-coarse diagnostic.}
To test whether fine patches alone are sufficient, we remove the coarse layer and reallocate the entire 1.0~KB budget to additional fine patches. As shown in~\cref{tab:no_coarse}, zero-filling missing regions performs poorly, and even adding a validity mask remains below our dual-resolution design. This confirms that a dense low-resolution semantic floor is more useful than simply maximizing the number of high-resolution patches under the same byte budget.
\begin{table}[h!]
\centering
\caption{\textbf{No-coarse diagnostic at 1.0~KB.}
Removing the coarse base layer degrades performance, especially when missing regions are zero-filled.}
\label{tab:no_coarse}
\scriptsize
\setlength{\tabcolsep}{4pt}
\begin{tabular}{l c c}
\toprule
\textbf{Setting} & \textbf{AP@0.5}$\uparrow$ & \textbf{AP@0.7}$\uparrow$ \\
\midrule
No-coarse + zero-fill & 0.360 & 0.290 \\
No-coarse + validity-mask & 0.663 & 0.540 \\
\textbf{Ours (DR)} & \textbf{0.706} & \textbf{0.588} \\
\bottomrule
\end{tabular}
\end{table}



\section{Conclusion}
We presented a dual-resolution collaborative perception framework that combines dense coarse BEV coverage with task-aware sparse refinement under kilobyte-scale budgets. The method preserves a standard dense fusion interface while enabling strict payload control and dynamic budget adaptation. Experiments on DAIR-V2X and OPV2V show strong accuracy--payload trade-offs across fusion modules, backbones, dynamic budgets, localization noise, and delay.

\noindent \textbf{Limitations and Future Work.}
One-shot broadcast does not remove cross-agent redundancy, so communication still grows with the number of collaborators. Future work will explore redundancy-aware coordination and task-specific refinement for downstream decision tasks.

\clearpage
\setcounter{page}{1}
\appendix
\pagebreak
\section*{Supplementary Material for Dense Coverage, Sparse Refinement: Byte-Constrained Cooperative Perception}
\addcontentsline{toc}{section}{Appendix}
\section{Hyperparameter}
\label{sec:app_hyperparameter}
\begin{table}[htb]
    \centering
    \caption{\textbf{Methodological Configuration.} Condensed hyperparameters for training, architecture, and the differentiable selector (Train $|$ Inference). Unless stated otherwise, dataset-dependent values are reported as DAIR-V2X / OPV2V.}
    \label{tab:comprehensive_hparams}
    \scriptsize
    \begin{tabular}{@{}lc@{}}
        \toprule
        \textbf{Parameter} & \textbf{Value / Setting} \\
        \midrule
        \multicolumn{2}{@{}c@{}}{\textit{Training Setup}} \\
        \midrule
        Hardware & 1$\times$ RTX 4090 \\
        Epochs & 30 / 40 \\
        Batch size & 4 / 2 \\
        Optimizer & AdamW \\
        LR (detector) & $1\times10^{-3}$ / $2\times10^{-3}$ \\
        LR (selector) & $2\times10^{-4}$ / $5\times10^{-4}$ \\
        LR (codebook) & $5\times10^{-4}$ / $1\times10^{-3}$ \\
        Warm-up epochs & 5 / 5 \\
        Warm-up LR & $2\times10^{-4}$ / $2\times10^{-4}$ \\
        Scheduler & CosineAnneal\\
        \midrule
        \multicolumn{2}{@{}c@{}}{\textit{Architecture \& Loss Weights}} \\
        \midrule
        BEV Channels / Codebook ($K, d$) & 256 / (64, 64) \\
        Selector Attn. (Dim / Heads / Fusion) & 256 / 4 / SplitAttn \\
        Patch Cell Size ($c$) / Metadata & 16 / 48-bit \\
        Joint Loss Weights ($\alpha_{\text{vq}}, \alpha_{\text{rd}}$) & 0.5, 2.0 \\
        Detection Loss Weights (Cls. / Reg.) & 1.0, 2.0 \\
        Benefit Sup. Config. ($|S|$, $w_{\text{task}}$, $\gamma$ , $\beta_{\text{floor}}$) & 256, 0.5, 0.5, 0.1 \\
        \midrule
        \multicolumn{2}{@{}c@{}}{\textit{Selector Dynamics (Train $\rightarrow$ Inference)}} \\
        \midrule
        Temperature ($\tau$) & $1.0 \rightarrow 0.2$ $|$ Greedy \\
        Budget Penalty ($\lambda$) & Init 0, EMA 0.9 $|$ N/A \\
        Sampling ($X/Y$)\textsuperscript{*} & 50/128 $\rightarrow$ 50/30 $|$ Top-\(N_{\text{fine}}\) \\
        Target Budget ($B_{target}$) & 3.0 KB $|$ Dynamic \\
        Base Stride ($S$) & 4 \\
        \bottomrule
        \multicolumn{2}{@{}p{\linewidth}@{}}{\vspace{2pt}\textsuperscript{*}\textit{Codec Stabilization:} $X/Y$ denotes selecting $X$ highest-utility patches and $Y$ random patches. High initial random exposure ($k_{\text{rand}}=128$) prevents early codebook collapse, gradually transitioning to selector-focused exploitation.}
    \end{tabular}
\end{table}
\noindent\textbf{Selector Architecture Details.} The Pyramid Window Attention module operates directly on the pooled $48 \times 128$ spatial grid. It comprises a transformer block with an input/hidden dimension of 256, 4 attention heads (with a per-head dimension of 64), and learnable 2D relative positional embeddings. It processes the feature map through parallel window sizes of $4\times4$, $8\times8$, and $16\times16$, and aggregates the multi-scale outputs using a channel-wise Split Attention fusion mechanism.
\section{Architectural Justification (Preliminary Studies)}
\label{sec:app_arch_just}
We conducted a systematic ablation study directly on the DAIR-V2X dataset to determine the optimal compression strategy. These findings, summarized below, fundamentally shaped our dual-resolution design.

\subsection{Global Downsampling degrades Precision}
\label{sec:downsampling}
We investigated simply reducing the spatial resolution of the transmitted feature map to save bandwidth. We conducted this analysis using CoBEVT integrated with SimVQ \cite{zhu_addressing_2024} and a codebook size of $K=64$. As shown in ~\cref{tab:spatial_ablation}, applying a global $2\times$ compression factor ($H/2, W/2$) reduced the payload by $4\times$ but caused a catastrophic drop in high-precision detection (AP@0.7 dropped from 0.52 to 0.39).
\textit{Design Decision:} Instead of degrading the entire scene, we adopted a \textbf{sparse fine layer} at full resolution to preserve object fidelity.

\begin{table}[h!]
\centering
\footnotesize
\caption{\textbf{Justification for Fine Layer.} Preliminary analysis on DAIR-V2X shows that global downsampling harms precision.}
\label{tab:spatial_ablation}
\begin{tabular}{l c cc c}
\toprule
Method & Factor & AP@0.5 & AP@0.7 \\
\midrule
Full Res & $1\times$ & \textbf{0.71} & \textbf{0.52}  \\
Global Downsample & $2\times$ & 0.65 & 0.39 \\
\bottomrule
\end{tabular}
\end{table}
\subsection{VQ methods to prevent Codebook Collapse}
\label{sec:coodebook_collapse}
\begin{figure}[htbp]
     \centering
     \begin{subfigure}[b]{0.48\linewidth}
         \centering
         \includegraphics[width=\textwidth]{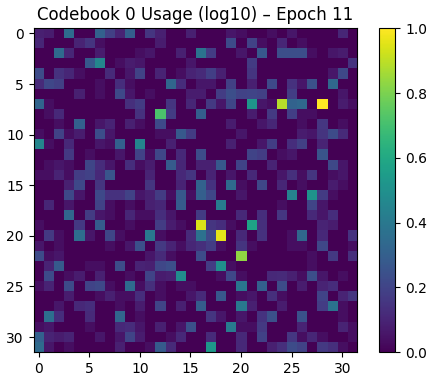}
         \caption{VQ Codebook Utilization}
         \label{fig:vq_util}
     \end{subfigure}
     \hfill 
     \begin{subfigure}[b]{0.48\linewidth}
         \centering
         \includegraphics[width=\textwidth]{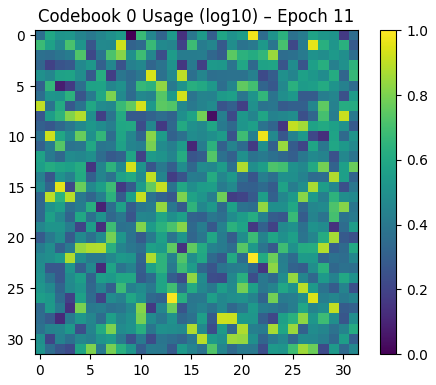}
         \caption{SimVQ Codebook Utilization}
         \label{fig:sim_util}
     \end{subfigure}

     \vspace{0.5em} 

     \begin{subfigure}[b]{0.48\linewidth}
         \centering
         \includegraphics[width=\textwidth]{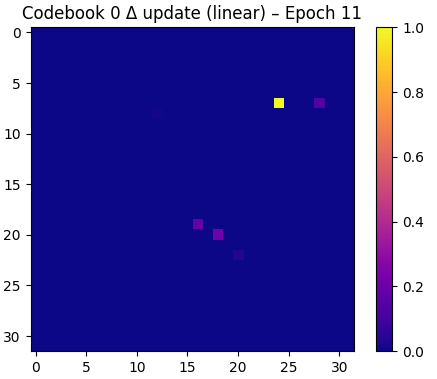}
         \caption{VQ Codebook Update}
         \label{fig:vq_update}
     \end{subfigure}
     \hfill 
     \begin{subfigure}[b]{0.48\linewidth}
         \centering
         \includegraphics[width=\textwidth]{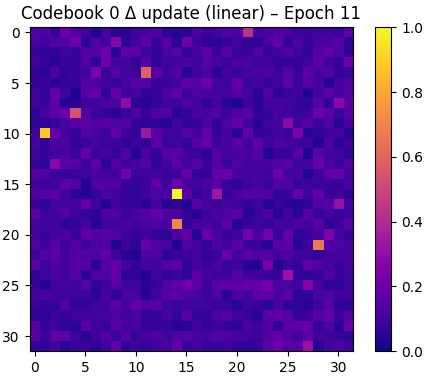}
         \caption{SimVQ Codebook Update}
         \label{fig:sim_update}
     \end{subfigure}

     \caption{Comparison of codebook dynamics at Epoch 11. (a) and (c) show that standard VQ collapses, with only a few vectors being utilized or updated. (b) and (d) demonstrate that SimVQ maintains a healthy, distributed codebook usage and consistent updates across the entire grid.}
     \label{fig:codebook_collapse}
\end{figure}

Standard VQ often suffers from codebook under-utilization (dead codes) under strict bandwidth budgets, which reduces bottleneck capacity and destabilizes end-to-end training~\cite{roy2018theory,sonderby2017continuous}. We therefore use SimVQ~\cite{zhu_addressing_2024}, which learns a reparameterization of the code space to improve codebook optimization and mitigate collapse in practice; empirically, it achieves a stronger rate-accuracy trade-off than Finite Scalar Quantization (FSQ) in our setting (Supp. Sec. B.3). 
Codebook collapse occurs when a large majority of the available codebook vectors are not selected during training, effectively reducing the model's capacity and leading to poor representation learning. To address this, we evaluated SimVQ by Zhu et al. \cite{zhu_addressing_2024}, which is designed to improve codebook utilization. We compared SimVQ against a standard VQ-VAE baseline \cite{oord_neural_2018} using CoBEVT on the OPV2V dataset with a codebook size of $K=1024$. To monitor the training dynamics, we used two primary metrics. Codebook usage is visualized as a heatmap showing how frequently each entry is selected, utilizing a $\log_{10}(x+1)$ transformation to ensure that low-frequency entries remain visible. The codebook update ($\Delta$) is visualized as the $L_2$ distance between a codeword at the current and previous epoch, defined as $\| c^{(t)} - c^{(t-1)} \|_2$, which indicates which specific vectors are actively being updated during the training process. As shown in \cref{fig:codebook_collapse}, the standard VQ approach suffers from severe collapse. By Epoch 11, only four codebook vectors receive any significant updates (\cref{fig:vq_update}), and the usage remains concentrated in a tiny fraction of the available slots (\cref{fig:vq_util}). This lack of diversity is further evidenced by a validation loss that rises steadily from the beginning of training. In contrast, SimVQ maintains high codebook utilization (\cref{fig:sim_util}) and ensures that a wide variety of vectors are actively updated throughout the training process (\cref{fig:sim_update}). 
\textit{Design Decision:} Based on these results, we use SimVQ for all subsequent experiments. 

\noindent\subsection{Learned Codebooks outperform Scalar Quantization}
We evaluate learned Vector Quantization (SimVQ~\cite{zhu_addressing_2024}) against Finite Scalar Quantization (FSQ)~\cite{mentzer_finite_2023} and the Autoencoder (AE) compression used in CoBEVT~\cite{xu_cobevt_2022}. FSQ uses a fixed, non-learnable grid to quantize latent dimensions independently, whereas SimVQ learns a re-parameterization of the code space that improves codebook optimization and mitigates dead-code behavior in practice. Architectural details are summarized in ~\cref{tab:model_configs}. ~\cref{tab:fsq_vs_vq} highlights the rate--accuracy gap: FSQ requires \(\sim\)24--49\,KB (\(\approx\)5--11\(\times\) higher payload) yet yields only a marginal gain in AP@0.5, while SimVQ operates at 4.61\,KB with competitive accuracy. SimVQ also outperforms the CoBEVT-VAE baseline while using \(\sim\)20\(\times\) less bandwidth. \textit{Design Decision:} We adopt learned codebooks (SimVQ) to maximize spectral efficiency under strict bandwidth constraints.
\begin{table}[ht]
\centering
\caption{\textbf{Model Configuration Comparison.} A detailed breakdown of architectural parameters for AE, SimVQ and FSQ.}
\label{tab:model_configs}
\footnotesize
\begin{tabular}{l cccc}
\toprule
Method & (VAE) & (SimVQ) & (FSQ v1) & (FSQ v2)\\
\midrule
Input Channels & 256 & 256 & 256 & 256 \\
Latent Channels & 4 & 64 & 32 & 16 \\
Compression Ratio & 64$\times$ & - & - & -\\
Codebook Size ($K$) & - & 64 & - & -\\
Num. Codebook & - & - & 16 & 16\\
Level & - & - & 4 & 4\\
\bottomrule
\end{tabular}
\end{table}
\begin{table}[h!]
\centering
\footnotesize
\caption{\textbf{Justification for Learned Codebooks.} Comparison of Learned VQ against FSQ and the Autoencoder compression shows that Learned VQ achieves the best balance between accuracy and bandwidth.}
\label{tab:fsq_vs_vq}
\begin{tabular}{l ccc }
\toprule
Method & AP@0.5$\uparrow$& AP@0.7$\uparrow$ & KB$\downarrow$ \\
\midrule
FSQ v1 & 0.71 & 0.58 & 49.15 \\
FSQ v2 & 0.73 & 0.57 & 24.58 \\
CoBEVT-VAE & 0.61 & 0.47 & 98.3 \\
\textbf{CoBEVT-SimVQ} & 0.71 & 0.52 & 4.61 \\
\bottomrule
\end{tabular}
\end{table}

\subsection{Latent Channel Dimension vs. Codebook Efficiency}
\label{sec:channel_ablation}
\cref{tab:channel_cb_size} summarizes experiments using CoBEVT with the SimVQ module to analyze the interplay between latent dimensionality and quantization performance. Our analysis highlights the relationship between the number of latent channels and overall system efficiency. Interestingly, reducing the latent channel dimension from the default feature depth of 256 down to 64 yields noticeable performance gains (AP@0.7 increases from 0.500 to 0.523) while simultaneously decreasing the VQ module's parameter count by nearly 90\%. \textit{Design Decision:} We hypothesize that this lower-dimensional bottleneck encourages the model to capture more compact, sparse semantic information, effectively regularizing the latent space for quantization. Consequently, we adopt a latent dimension of 64 to minimize both the computational footprint and the risk of codebook under-utilization.
\begin{table}[h!]
\centering
\footnotesize
\caption{\textbf{Impact of Latent Dimension.} Reducing latent channels improves accuracy and efficiency.}
\label{tab:channel_cb_size}
\resizebox{\linewidth}{!}{%
\begin{tabular}{l c c c c}
\toprule
$K$ & Latent Channels & AP@0.7$\uparrow$ & Inference time (ms) & Params (M)  \\
\midrule
64 & 64 & 0.523 & 19.6 & 0.596 \\
64 & 128 & 0.508 & 21.6 & 1.79 \\
64 & 256 & 0.500 & 24.0 & 5.97 \\
\bottomrule
\end{tabular}%
}
\end{table}

\subsection{VQ vs. Residual VQ}
\label{sec:vq_vs_rvq}
Unlike standard Vector Quantization (VQ), which maps a latent vector to a single index, Residual Vector Quantization (RVQ \cite{zeghidour_soundstream_2021}) employs a cascade of multiple quantizers. The first stage provides a coarse approximation of the feature, while each subsequent stage iteratively refines the residual error from the previous step. The evaluation shows that while Residual Vector Quantization (RVQ) significantly improves detection accuracy, it inherently increases the communication overhead. A key advantage of the residual approach is that it allows the model to be deployed using a variable number of quantizers without the need for retraining. For instance, the number of quantization steps can be reduced to a single stage at runtime to accommodate lower bandwidth requirements or limited computational resources. However, our evaluation shows that using only the first quantization step of a multi-stage RVQ model is less performant than a dedicated VQ model trained for that specific bit budget, as illustrated in ~\cref{tab:svq_rvq_payload}. \textit{Design Decision:} Given the strict bandwidth constraints of V2X communication, we prioritize the efficiency of standard VQ over the runtime adaptability of RVQ.

\begin{table}[h!]
\centering
\footnotesize
\caption{VQ vs.\ Residual VQ. While RVQ benefits from additional stages, 
standard VQ achieves better accuracy-per-bit compared to the first RVQ stage.}
\label{tab:svq_rvq_payload}
\setlength{\tabcolsep}{8pt}
\begin{tabular}{l l c c c}
\toprule
Model & Stages & Payload (KB)$\downarrow$ & AP@0.5$\uparrow$ & AP@0.7$\uparrow$ \\
\midrule

\multicolumn{5}{l}{\textbf{Codebook Size 64}} \\
VQ  & 1 & 4.61  & 0.71 & 0.52 \\
RVQ & 1 & 4.61  & 0.71 & 0.51 \\
RVQ & 3 & 13.82 & 0.70 & 0.54 \\
\midrule

\multicolumn{5}{l}{\textbf{Codebook Size 128}} \\
VQ  & 1 & 5.38  & 0.71 & 0.51 \\
RVQ & 1 & 5.38  & 0.70 & 0.50 \\
RVQ & 3 & 16.13 & 0.71 & 0.54 \\
\midrule

\multicolumn{5}{l}{\textbf{Codebook Size 256}} \\
VQ  & 1 & 6.14  & 0.71 & 0.51 \\
RVQ & 1 & 6.14  & 0.71 & 0.50 \\
RVQ & 3 & 18.43 & 0.72 & 0.55 \\
\midrule

\multicolumn{5}{l}{\textbf{Codebook Size 512}} \\
VQ  & 1 & 6.91  & 0.72 & 0.53 \\
RVQ & 1 & 6.91  & 0.71 & 0.52 \\
RVQ & 3 & 20.74 & 0.71 & 0.54 \\
\midrule

\multicolumn{5}{l}{\textbf{Codebook Size 1024}} \\
VQ  & 1 & 7.68  & 0.71 & 0.52 \\
RVQ & 1 & 7.68  & 0.70 & 0.51 \\
RVQ & 3 & 23.04 & 0.72 & 0.55 \\

\bottomrule
\end{tabular}
\end{table}

\subsection{End-to-End Optimization outperforms Staged Training}
\label{sec:app_training_ablations}
We investigated whether a curriculum learning approach could improve stability by decoupling feature learning from quantization. We compared a staged approach, where the backbone is pre-trained before activating and fine-tuning the compression module—against two joint training strategies: one where the VQ module is initialized with pre-trained weights and kept frozen, and the standard approach where the entire pipeline is optimized jointly from scratch. Contrary to expectations, complex schedules yielded suboptimal results. As shown in ~\cref{tab:train_strategy}, both the staged fine-tuning strategy (AP@0.7 of 0.49) and the frozen codebook approach (AP@0.7 of 0.49) failed to match the performance of joint optimization from scratch (AP@0.7 of 0.52).

\textit{Design Decision:} We hypothesize that freezing or pre-training the codebook limits the encoder's ability to co-adapt to quantization errors. Consequently, we adopt a fully \textbf{end-to-end training strategy}, allowing the perception backbone and compression module to optimize simultaneously.
\begin{table}[h!]
\centering
\footnotesize
\caption{\textbf{Impact of Training Strategy.} Joint optimization from scratch outperforms both staged fine-tuning and frozen codebook strategies.}
\label{tab:train_strategy}
\begin{tabular}{l c c}
\toprule
Strategy & AP@0.5$\uparrow$ & AP@0.7$\uparrow$ \\
\midrule
Staged (Fine-Tuned) & 0.70 & 0.49 \\
Joint Training (Frozen Codebook) & 0.69 & 0.49 \\
\textbf{End-to-End (Joint from Scratch)} & \textbf{0.71} & \textbf{0.52} \\
\bottomrule
\end{tabular}
\end{table}
\subsection{Robustness to Prior Formulation}
\label{sec:prior_formulation}
\paragraph{Selection Agreement.}
In the main paper, the cell-level task prior $\psi_{h,w}$ is derived from normalized feature magnitude. 
To evaluate sensitivity to this choice, we replace $\psi_{h,w}$ at inference with a detector-derived confidence prior obtained from the classification head (PSM). 
All other components, including distortion prediction $\delta$ and greedy ranked selection, remain unchanged.

\begin{figure}[t]
\centering
\begin{subfigure}{0.49\linewidth}
    \centering
    \includegraphics[width=\linewidth]{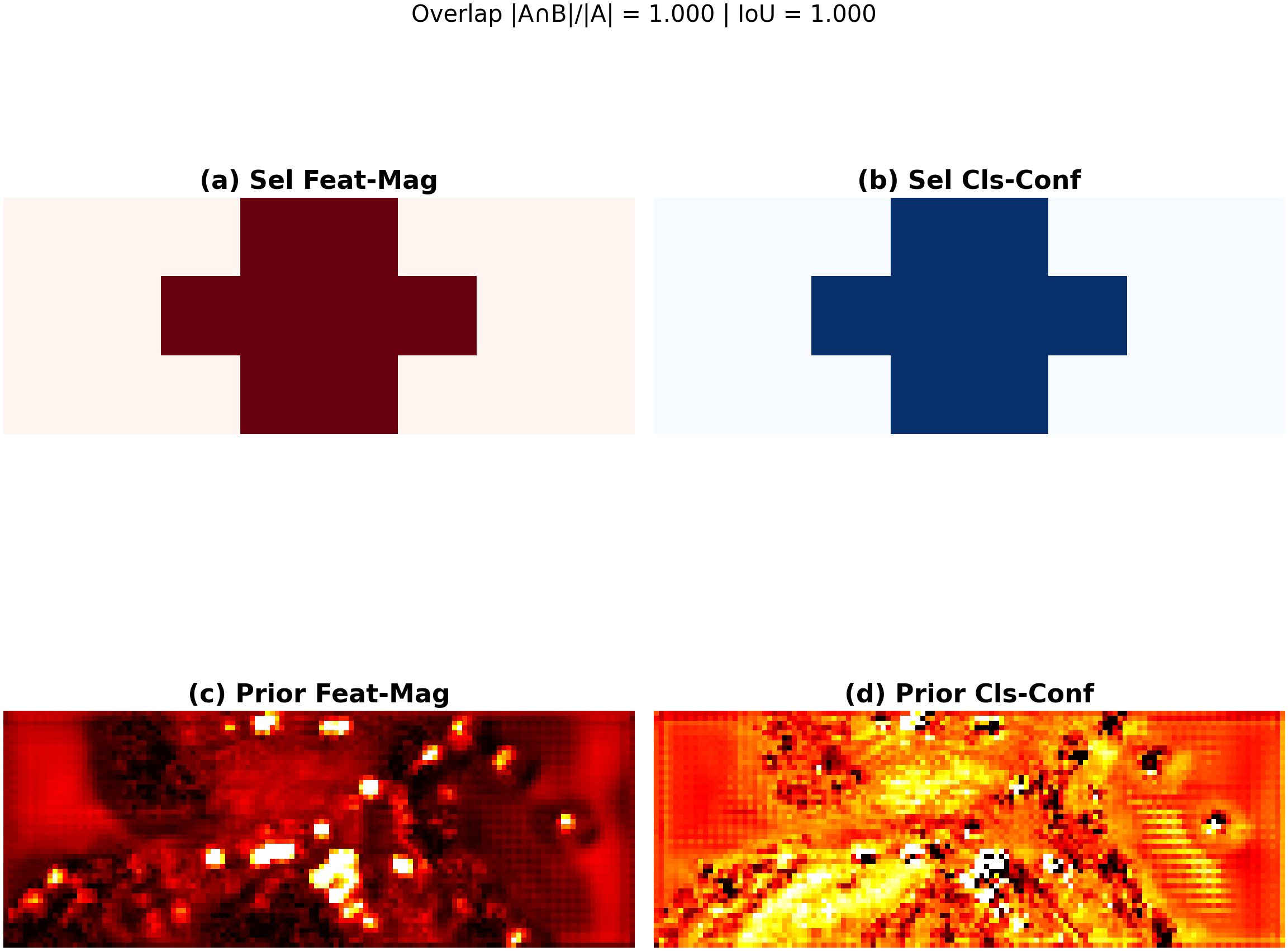}
    \caption{2KB}
\end{subfigure}
\hfill
\begin{subfigure}{0.49\linewidth}
    \centering
    \includegraphics[width=\linewidth]{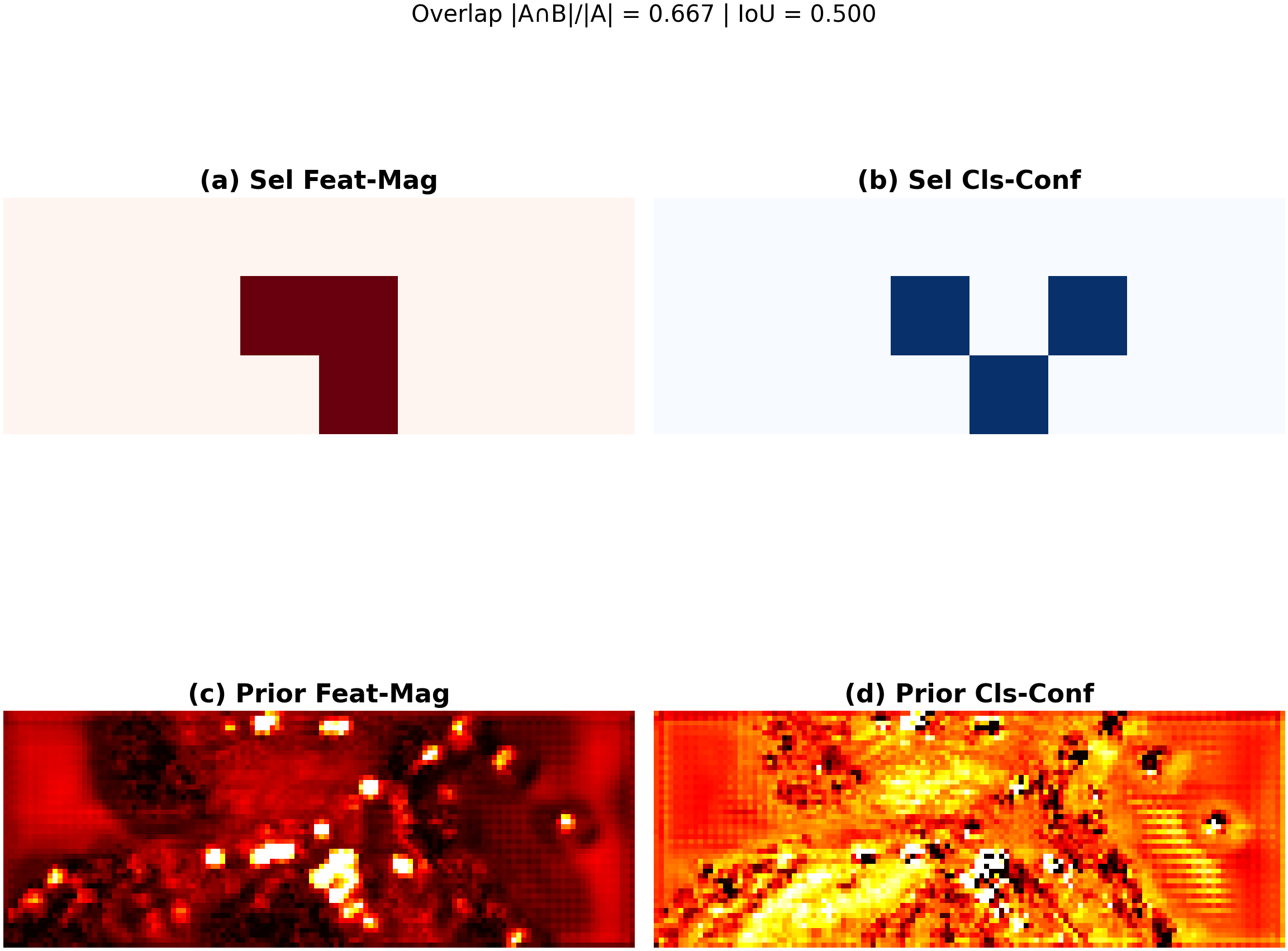}
    \caption{1KB}
\end{subfigure}
\caption{Selection robustness under different priors. Top: binary cell selection masks; bottom: corresponding prior heatmaps.}
\label{fig:prior_robustness}
\end{figure}

Under moderate bandwidth (e.g., 2KB), the selection masks are identical (overlap = 1.000, IoU = 1.000) in several scenes, despite clear visual differences in the prior heatmaps. 
At stricter budgets (e.g., 1KB), agreement remains substantial (overlap = 0.667, IoU = 0.500), where discrete truncation effects amplify small score differences. The selections and heatmaps are illustrated in Figure~\ref{fig:prior_robustness}.

\paragraph{Impact on Detection.}
Across all evaluated budgets, detection performance remains unchanged when replacing the prior. 
This indicates that ranking is primarily governed by the learned distortion-aware term $\sigma(\delta_{h,w})$, while $\psi_{h,w}$ serves as a stabilizing modulation rather than the dominant selection driver.

\paragraph{Discussion.}
These results demonstrate that the magnitude-based prior is not a brittle heuristic. 
Instead, the overall selection mechanism is structurally robust to prior formulation, suggesting that the dominant factor in selection is the learned rate-aware utility prediction.
\section{Selector Architecture Trade-Off}
To justify our use of the Pyramid Window Attention module, we compare its rate-distortion ranking performance against a lightweight Dilated CNN baseline. Both architectures are trained with the same task-aware supervision ($\Psi$). As shown in \cref{tab:selector_arch_supp}, although the Dilated CNN is computationally efficient, requiring only 0.39\,M parameters and 9.54\,GFLOPs, its local receptive field limits its ability to globally rank competing spatial patches.

This limitation becomes particularly apparent under severe bandwidth constraints. At a strict 1.0\,KB budget, Pyramid Attention outperforms the Dilated CNN by \textbf{2.0 percentage points} at AP@0.5 (0.706 vs.\ 0.686) and \textbf{3.3 percentage points} at AP@0.7 (0.588 vs.\ 0.555). This gap indicates that capturing global, multi-scale context is critical when most of the feature map must be discarded. At a more relaxed 2.0\,KB budget, the gap narrows at AP@0.5 (0.736 vs.\ 0.725), but Pyramid Attention still provides better fine-grained localization at AP@0.7 (0.604 vs.\ 0.580). We therefore adopt Pyramid Attention, concluding that its additional compute cost is a worthwhile trade-off for improved spectral efficiency in the kilobyte regime.
\begin{table}[h!]
\scriptsize
\centering
\caption{\textbf{Selector architecture trade-off.} Pyramid Attention improves ranking under strict budgets compared to a local Dilated CNN (both with task-aware supervision $\Psi$).}
\label{tab:selector_arch_supp}
\setlength{\tabcolsep}{4pt}
\begin{tabular}{l l c c c c}
\toprule
\textbf{Budget} & \textbf{Architecture} & \textbf{AP@0.5} & \textbf{AP@0.7} & \textbf{Params (M)} & \textbf{GFLOPs} \\
\midrule
1.0 KB & Dilated CNN & 0.686 & 0.555 & 0.39 & 9.54 \\
1.0 KB & \textbf{Pyramid Attn.} & \textbf{0.706} & \textbf{0.588} & 1.27 & 28.84 \\
\midrule
2.0 KB & Dilated CNN & 0.725 & 0.580 & 0.39 & 9.54 \\
2.0 KB & \textbf{Pyramid Attn.} & \textbf{0.736} & \textbf{0.604} & 1.27 & 28.84 \\
\bottomrule
\end{tabular}
\end{table}
\section{Detailed Payload Composition}
\label{sec:appendix_payload_analysis}

To validate the feasibility of our reported operating point, we provide a byte-level breakdown demonstrating strict adherence to the bandwidth budget.
We report the actual payload transmitted per non-ego agent in KB, computed directly from the bitstreams produced by our implementation (\(1~\mathrm{KB}=1{,}000\) bytes). We do not apply entropy coding. Let \(K\) be the codebook size, \(n_{\text{coarse}}\) the total number of transmitted coarse VQ indices, \(n_{\text{fine}}\) the number of VQ indices per selected fine patch, and \(N_{\text{fine}}\) the number of selected fine patches. The total payload in bits is:
\begin{equation}
B_{\text{total}}(i) =
\bigl(n_{\text{coarse}} + N_{\text{fine}} \cdot n_{\text{fine}}\bigr)\log_2(K)
+ N_{\text{fine}} \cdot B_{\text{header}},
\end{equation}
where \(B_{\text{header}}\) is the per-patch metadata cost. Crucially, our selector strictly respects the bandwidth cap \(B_{\text{target}}\). Below, we decompose the 2.0~KB operating point on DAIR-V2X~\cite{yu_dair-v2x_2022}.

\subsection{System Parameters}
\begin{itemize}
    \item \textbf{BEV Feature Map \(\mathbf{F}\):} \(48 \times 128\) features, resulting from a \(4\times\) backbone stride on the \(192 \times 512\) voxel grid.
    \item \textbf{Fine Cell Size:} \(16 \times 16\) BEV feature locations.
    \item \textbf{Codebook (\(K\)):} 64 entries, corresponding to 6 bits per token.
    \item \textbf{Header (\(B_{\text{header}}\)):} 48 bits per fine patch, conservatively covering spatial coordinates, CRC-16, and framing.
    \item \textbf{Budget Constraint (\(B_{\text{target}}\)):} 2.0~KB, i.e., 16{,}000 bits.
\end{itemize}

\subsection{Strict Budget Enforcement}
\noindent \textbf{1. Fixed overhead: coarse base layer.}
The coarse base layer is spatially downsampled by \(S_{\text{base}}=4\) in both dimensions, yielding a \(12 \times 32\) grid of coarse code indices from the valid \(48 \times 128\) feature area:
\begin{equation}
    B_{\text{coarse}} = (12 \times 32) \times 6 = 2{,}304 \text{ bits} \approx 0.288~\text{KB}.
\end{equation}
This leaves \(16{,}000 - 2{,}304 = 13{,}696\) bits for fine refinement.

\noindent \textbf{2. Per-patch fine cost.}
Each selected fine patch transmits its full-resolution \(16\times16\) VQ tokens plus the metadata header:
\begin{equation}
    B_{\text{patch}} = (256 \times 6) + 48 = 1{,}584 \text{ bits/patch}.
\end{equation}

\noindent \textbf{3. Maximum fine-patch count.}
The selector fills the remaining budget with the highest-ranked fine patches. The maximum number of selected fine patches is:
\begin{equation}
    N_{\text{fine}} =
    \left\lfloor \frac{13{,}696}{1{,}584} \right\rfloor
    = 8.
\end{equation}
A ninth fine patch would require \(9 \times 1{,}584 = 14{,}256\) bits, exceeding the available 13{,}696 bits. Therefore, the allocator selects eight fine patches.

\subsection{Total Payload Summary}
Table~\ref{tab:bitstream_composition} details the final bitstream for DAIR-V2X. The total payload is 1.872~KB, or 14{,}976 bits, satisfying the strict 2.0~KB constraint.

\begin{table}[h]
    \centering
    \caption{\textbf{Bitstream composition under a 2.0\,KB budget (DAIR-V2X).} 
    Under a 2.0\,KB cap, the system transmits the coarse floor and the top-8 highest-utility refinement patches. The 48-bit header includes the spatial coordinate required to reconstruct $M_i$ at the receiver.}
    \label{tab:bitstream_composition}
    \resizebox{\linewidth}{!}{%
        \begin{tabular}{@{}llrr@{}}
            \toprule
            \textbf{Component} & \textbf{Count} & \textbf{Size (Bits)} & \textbf{Size (KB)} \\
            \midrule
            1. Coarse Base Layer & 1 (Full $12 \times 32$ Grid) & 2,304 & 0.288 \\
            2. Fine Patches (VQ Indices) & 8 patches & 12,288 & 1.536 \\
            3. Per-Patch Headers & 8 headers & 384 & 0.048 \\
            \midrule
            \textbf{Total Payload} & & \textbf{14,976} & \textbf{1.872} \\
            \bottomrule
        \end{tabular}%
    }
\end{table}
\subsection{Baseline Accounting Used in the Main Paper.} All KB results in the main paper use the decodable accounting (including indices where required for sparse representations). Unless noted, all KB values denote the total transmitted payload per non-ego agent per fusion step (computed from decodable messages); for request–response protocols, we conservatively include both ego$\to$agent and agent$\to$ego messages.
\paragraph{\textbf{Where2comm (mask-based sparse FP32)}.}
Mask-based approaches transmit only the spatial positions selected by a binary communication mask.
Let $\mathrm{Mask}_i \in \{0,1\}^{H\times W}$ denote the selection mask for non-ego agent~$i$,
$K_i = \sum_{u,v} \mathrm{Mask}_i(u,v)$ the number of selected cells,
$C$ the feature channel dimension, and $\mathrm{sizeof(dtype)}$ the bytes per element
(4 for FP32).

\paragraph{Overhead assumption.}
The receiver does not know $\mathrm{Mask}_i$ a priori; it must be transmitted.
Following a decodable accounting without entropy coding, we assume the sender transmits the
full binary mask as a dense bitfield ($H \!\times\! W$ bits).\footnote{%
  An alternative is to transmit $K_i$ position indices, which (with optimal bit packing) costs
  $K_i \lceil \log_2(HW) \rceil$ bits. Index coding is cheaper when
  $K_i < HW / \lceil\log_2(HW)\rceil$; otherwise a dense bitfield is cheaper.
  We adopt dense bitfields as a simple, decodable accounting without entropy coding.}
The total communication cost per frame is therefore:
\begin{equation}
\resizebox{\linewidth}{!}{%
  $\mathrm{Comm}_{\mathrm{KB}}^{\text{Where2comm}} = \frac{1}{1000}\sum_{i\in\mathcal{N}}\!\Bigl[
      \underbrace{K_i \cdot C \cdot \mathrm{sizeof(dtype)}}_{\text{payload (feature values)}}
      \;+\;
      \underbrace{\frac{H \times W}{8}}_{\text{mask overhead (bytes)}}
    \Bigr].$
}
\end{equation}
For multi-scale architectures with $L$ feature levels of spatial size
$(H_\ell, W_\ell)$, the mask overhead generalises to
$\sum_{\ell=1}^{L} H_\ell W_\ell / 8$ bytes per agent.

\paragraph{\textbf{EffiComm (two-stage mask-based sparse FP32)}.}
EffiComm applies two cascaded spatial selections:
(i)~a confidence-based binary mask $\mathrm{Mask}_i^{(1)} \in \{0,1\}^{H\times W}$
(identical to Where2comm's Communication mask), followed by
(ii)~an adaptive grid reduction that retains a GNN-predicted fraction
$\rho_i$ of the remaining cells via a second top-$k$ mask
$\mathrm{Mask}_i^{(2)}$.
The effective selection is
$\mathrm{Mask}_i = \mathrm{Mask}_i^{(1)} \odot \mathrm{Mask}_i^{(2)}$
with $K_i = \sum_{u,v}\mathrm{Mask}_i(u,v)$ surviving cells.

\paragraph{Overhead assumption.}
The receiver does not know $\mathrm{Mask}_i$ a priori; it must be transmitted.
We transmit a single \emph{composite} binary mask as a dense bitfield
($H\!\times\!W$ bits) per agent; the receiver does not need to reconstruct
each selection stage separately.\footnote{%
  An alternative is index coding, which (with optimal bit packing) costs
  $K_i\lceil\log_2(HW)\rceil$ bits. Index coding is cheaper when
  $K_i < HW/\lceil\log_2(HW)\rceil$. We adopt dense bitfields as a simple,
  decodable accounting without entropy coding.}

\begin{equation}
\resizebox{\linewidth}{!}{%
  $\mathrm{Comm}_{\mathrm{KB}}^{\text{EffiComm}} = \frac{1}{1000}\sum_{i\in\mathcal{N}} \left[ \underbrace{K_i \cdot C \cdot \mathrm{sizeof(dtype)}}_{\text{payload}} + \underbrace{\frac{H \times W}{8}}_{\text{mask overhead}} \right]$%
}
\end{equation}
\paragraph{\textbf{ERMVP (token-based).}}
ERMVP selects $K_i$ tokens per agent via top-$k$ scoring followed by DPC-KNN clustering,
and transmits each token's feature vector together with its position on the $H\!\times\!W$ grid.

\paragraph{Overhead assumption.}
Because $K_i \ll HW$ in the typical operating regime
(e.g., $\texttt{topk\_ratio}\approx 0.05$--$0.25$), it is cheaper to send
\emph{position indices} rather than a full $H\!\times\!W$ bitfield mask.\footnote{%
  Break-even: index coding is cheaper when $K_i < HW / \lceil\log_2(HW)\rceil$.
  For $(H,W)=(60,180)$, $HW=10{,}800$ and $\lceil\log_2(HW)\rceil=14$, so the threshold is
  $10{,}800/14\approx 771$ tokens.}

\paragraph{Empirical cost (as implemented).}
In the released implementation, each token carries a \emph{single} flattened grid index
$\in[0,HW)$ stored as \texttt{torch.int64} (8 bytes). Thus $\mathrm{sizeof(pos\_dtype)}=8$ and
the per-agent payload is:
\begin{equation}
\resizebox{\linewidth}{!}{%
  $\mathrm{Comm}_{\mathrm{KB}}^{\mathrm{emp}} = \frac{1}{1000}\sum_{i\in\mathcal{N}} K_i \Bigl( \underbrace{C \cdot \mathrm{sizeof(dtype)}}_{\text{payload}} \;+\; \underbrace{8}_{\text{position index (\texttt{int64})}} \Bigr).$
}
\end{equation}

\paragraph{Theoretical lower bound.}
An entropy-optimal representation would bit-pack the flattened index using
$\lceil\log_2(HW)\rceil$ bits per token:
\begin{equation}
  \mathrm{Comm}_{\mathrm{KB}}^{\mathrm{theory}}
  = \frac{1}{8000}\sum_{i\in\mathcal{N}} K_i
    \Bigl(
      \underbrace{C \cdot 32}_{\text{payload (bits)}}
      \;+\;
      \underbrace{\lceil\log_2(HW)\rceil}_{\text{index overhead (bits)}}
    \Bigr).
\end{equation}
 
\paragraph{\textbf{MMCooper}.} MMCooper transmits (i) sparse masked feature positions and (ii) filtered bounding boxes. Let $\overline{m}$ be the average number of masked feature positions (each with $C=64$ channels) and $\overline{n}$ the average number of bounding boxes (each with 7 parameters) across batch size $B$: \begin{equation} \overline{m} = \frac{1}{B} \sum_{b=1}^{B} \sum_{u,v} \mathbf{1}[\mathrm{Mask}^{\text{feat}}_{b}(u,v) > 0], \quad \overline{n} = \frac{1}{B} \sum_{b=1}^{B} N^{\text{box}}_{b}. \end{equation} The original implementation reports a log-scale proxy: \begin{equation} \mathrm{Comm}_{\log_2\text{bytes}} = \log_2 \left( \left( C \cdot \overline{m} + 7 \cdot \overline{n} \right) \cdot 4 \right) \end{equation} where $C=64$ is the feature channel count and $4$ bytes converts FP32 elements. For direct KB comparison, we convert to bytes. The lower-bound (features + boxes only) is: \begin{equation} \mathrm{Comm}_{\mathrm{KB}}^{\text{lb}} = \frac{(C \cdot \overline{m} + 7 \cdot \overline{n}) \cdot 4}{1000} \end{equation} For decodable transmission, position indices must be included. Using 4 bytes per position (flattened int32 index or two int16 coordinates): \begin{equation} \mathrm{Comm}_{\mathrm{KB}}^{\text{idx}} = \mathrm{Comm}_{\mathrm{KB}}^{\text{lb}} + \frac{\overline{m} \cdot 4}{1000} \end{equation}
\paragraph{Instantiation:} We report $\mathrm{Comm}_{\mathrm{KB}}^{\text{idx}}$ (decodable) in main plots. Index dtype: \texttt{int32} (4 bytes per position). We treat each transmitted box as 7 FP32 values; if the implementation transmits additional fields (e.g., score, class ID), the payload increases linearly and can be added straightforwardly. 
\paragraph{\textbf{CoSDH (supply-demand aware intermediate-late hybrid)}.}
CoSDH performs communication-efficient collaboration by (i) selecting sparse collaboration regions via a supply--demand mask, (ii) transmitting \emph{sparse} multi-scale intermediate features with autoencoder-based channel compression, and (iii) optionally transmitting detection results for confidence-aware late fusion.

\paragraph{Supply-demand selection.}
Each agent $i$ produces a binary demand mask $D_i\in\{0,1\}^{H\times W}$ indicating where it needs information, and each collaborating agent $j$ produces a binary supply mask $S_j\in\{0,1\}^{H\times W}$ from its detection confidence map. The selection mask is
\begin{equation}
M_{j\to i} = D_i \odot S_j \in \{0,1\}^{H\times W},
\end{equation}
which is downsampled per scale and applied to multi-scale BEV features $\{F_j^{(l)}\}_{l=1}^{L}$ to obtain sparse features $\{Z_{j\to i}^{(l)}\}_{l=1}^{L}$. During communication, only the \emph{non-zero parts} and their \emph{corresponding coordinates} are transmitted~\cite{xu_cosdh_2025}.

\paragraph{Intermediate feature message (per agent $j\to i$).}
For each scale $l\in\{1,\dots,L\}$, let $H_l\times W_l$ and $C_l$ denote the spatial and channel dimensions of the $l$-th feature map, and let $K_{j\to i,l}$ be the number of selected positions at that scale.
CoSDH compresses sparse features along the channel dimension using a per-scale autoencoder with compression ratio $c_0$. Before transmission, compressed features are converted from float32 to float16 \cite{xu_cosdh_2025}.
A decodable cost for the intermediate message is:
\begin{equation}
\resizebox{\linewidth}{!}{%
  $\mathrm{Bytes}^{\mathrm{inter}}_{j\to i} = \sum_{l=1}^{L} K_{j\to i,l} \left( \underbrace{\frac{C_l}{c_0} \cdot 2}_{\text{FP16 feature values}} + \underbrace{\mathrm{Bytes}_{\mathrm{idx}}(H_lW_l)}_{\text{coordinates}} \right).$
}
\end{equation}
where $\mathrm{Bytes}_{\mathrm{idx}}(H_lW_l)$ is the coordinate overhead per selected position (e.g., $\lceil\log_2(H_lW_l)\rceil/8$ bytes with optimal bit packing, or a fixed integer dtype).

\paragraph{Late fusion message (optional, per agent $j\to i$).}
CoSDH optionally transmits detection results for confidence-aware late fusion; dense predictions before NMS are transmitted~\cite{xu_cosdh_2025}.
Let $\mathrm{Bytes}^{\mathrm{late}}_{j\to i}$ denote this cost.

\paragraph{Total response-only cost.}
The total decodable payload received by ego agent $i$ from collaborators is:
\begin{equation}
\resizebox{\linewidth}{!}{%
  $\mathrm{Comm}_{\mathrm{KB}}^{\mathrm{CoSDH}} = \frac{1}{1000}\sum_{j\neq i} \left( \mathrm{Bytes}^{\mathrm{inter}}_{j\to i} + \mathbf{1}[\text{late fusion}] \cdot \mathrm{Bytes}^{\mathrm{late}}_{j\to i} \right).$
}
\end{equation}

\noindent\paragraph{Demand-mask note.}
The demand mask $D_i$ is required to form $M_{j\to i}$. If explicitly broadcast, a bit-packed cost is $HW/8$ bytes per frame.

\paragraph{\textbf{InfoCom.}}
InfoCom transmits two components per non-ego agent: (i) an information-aware latent feature $\mathbf{E} \in \mathbb{R}^{D}$ from the IB encoder (FP32), and (ii) a sparse quantized mask $M^q$ containing only the $K = \lfloor \alpha \cdot H \cdot W \rfloor$ retained positions.
The paper's Eq.~(11) gives a lower bound (latent + mask values only):
\begin{equation}
\mathrm{Bits}_{\text{paper}} = D \cdot 32 + K \cdot b.
\end{equation}
For decodable transmission, position indices must be included. We use bit-packed indices with $\lceil \log_2(HW) \rceil$ bits each, which is more favorable than storing indices as \texttt{int16}/\texttt{int32}:
\begin{equation}
\mathrm{Bits}_{\text{idx}} = K \cdot \lceil \log_2(H \cdot W) \rceil.
\end{equation}
Total decodable payload per agent:
\begin{equation}
\mathrm{Bits}_{\text{agent}} = D \cdot 32 + K \cdot b + K \cdot \lceil \log_2(H \cdot W) \rceil.
\end{equation}
For $N_{\mathrm{cav}}$ non-ego agents:
\begin{equation}
\mathrm{Comm}_{\mathrm{KB}} = \frac{N_{\mathrm{cav}} \cdot \mathrm{Bits}_{\text{agent}}}{8 \times 1000}.
\end{equation}

\noindent\textbf{Instantiation used in our evaluation:}
\begin{itemize}
    \item Latent dimension: $D=256$ (FP32, 32 bits)
    \item Mask quantization: $b=4$ bits per retained position
    \item Mask resolution: $H=48, W=176$ (OPV2V spatial range)
    \item Retention ratio: $\alpha=0.1$ (at convergence), yielding $K=\lfloor 0.1 \cdot 48 \cdot 176 \rfloor = 844$
    \item Index bits: Since $H \times W = 48 \times 176 = 8{,}448$, each retained position requires $\lceil \log_2(8{,}448) \rceil = 14$ bits under bit-packed indexing.
\end{itemize}

\begin{align*}
\mathrm{Bits}_{\mathrm{latent}} &= 256 \times 32 = 8{,}192 \text{ bits} = 1.024 \text{ KB} \\
\mathrm{Bits}_{\mathrm{mask}} &= 844 \times 4 = 3{,}376 \text{ bits} = 0.422 \text{ KB} \\
\mathrm{Bits}_{\mathrm{idx}}  &= 844 \times 14 = 11{,}816 \text{ bits} = 1.477 \text{ KB} \\
\mathrm{Total}_{\text{decodable}} &\approx 2.923 \text{ KB per non-ego agent}
\end{align*}

\noindent\textbf{Dense-mask alternative.}
If instead a dense $H \times W$ mask quantized to $b$ bits is transmitted (no indices), the per-agent cost is:
\begin{equation}
\resizebox{\linewidth}{!}{%
  $\mathrm{Comm}^{\text{dense}}_{\mathrm{KB}} = \frac{D \cdot 32 + H \cdot W \cdot b}{8 \times 1000} \approx 5.248 \text{ KB per non-ego agent}.$%
}
\end{equation}

\noindent\textbf{Note:}
We report $\mathrm{Comm}_{\mathrm{KB}}$ (decodable, with bit-packed indices) in main plots/tables.
\section{Sensitivity Analysis}
\begin{figure}[h!]
    \centering
    \includegraphics[width=0.8\columnwidth]{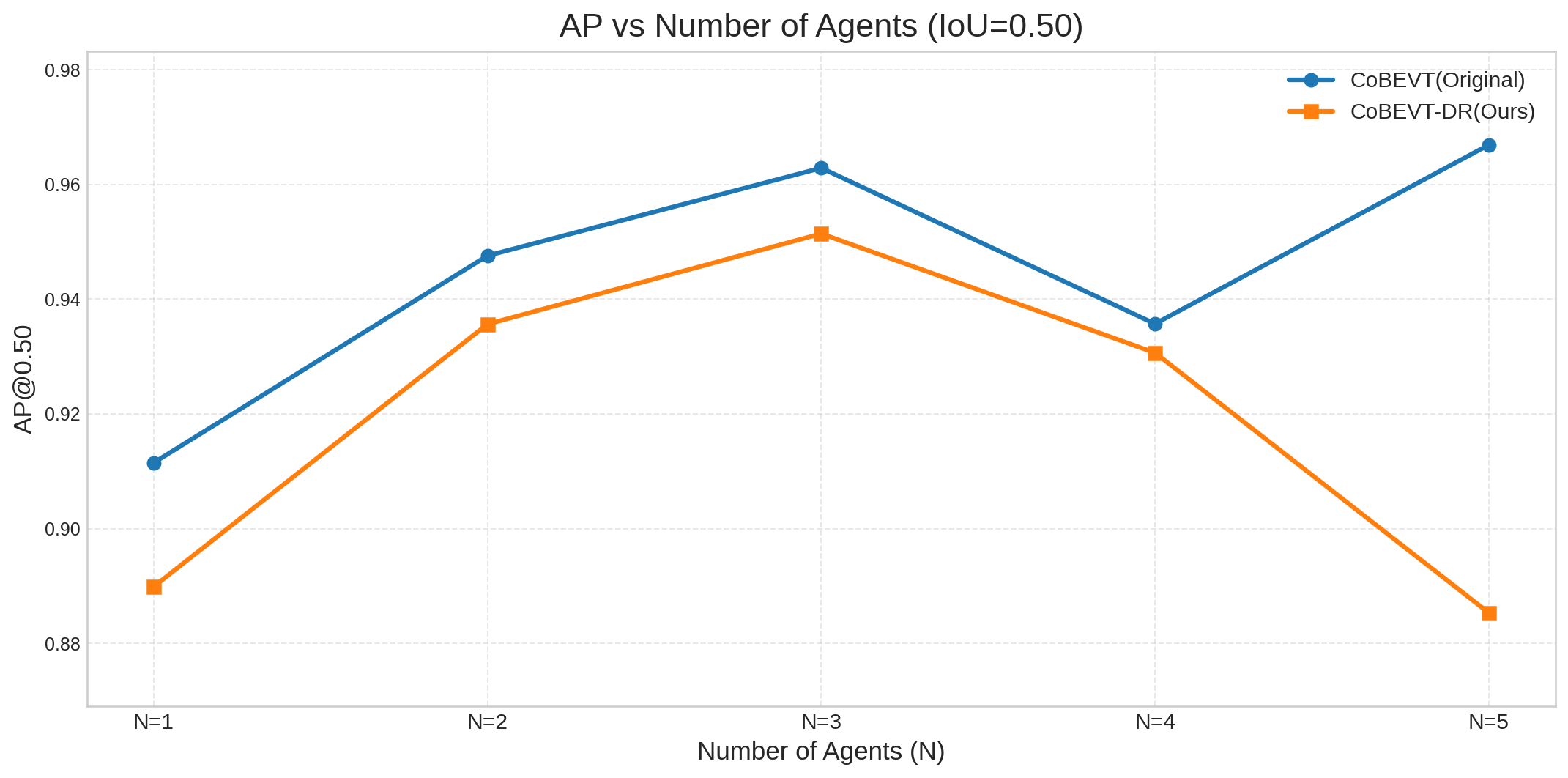}
\caption{\textbf{Scaling with number of agents under fixed budgets (OPV2V).}
AP@0.5 versus non-ego agent count \(N\) at a 2.0\,KB/agent cap for CoBEVT-DR.
Dense CoBEVT uses \(\sim\)6.3\,MB/agent (\(\sim\)31.4\,MB total network payload at \(N{=}5\)), whereas CoBEVT-DR uses \(\sim\)2.0\,KB/agent (\(\sim\)10\,KB total at \(N{=}5\)).}
    \label{fig:sens_analys}
\end{figure}

At a fixed per-agent budget (2.0 KB/agent), CoBEVT-DR improves from $N=1$ to $N=3$ and remains competitive at $N=4$, while performance drops at $N=5$ due to accumulated quantization artifacts from multiple highly-compressed sources, as illustrated in ~\cref{fig:sens_analys}. Importantly, matching the dense baseline at $N=5$ requires an impractical aggregate payload of ~31.4 MB per frame, whereas CoBEVT-DR uses only ~10 KB across all 5 agents. Since typical V2X collaboration most often involves $N \in \{2,3\}$, the proposed method scales favorably in the common operating regime.

\section{System Overhead and Real-Time Feasibility}
To validate deployment feasibility, we benchmark the wall-clock inference runtime on a single NVIDIA RTX 4090. We report the average per-frame latency for the encoder, selector, and fusion modules.

\begin{table}[h!] 
\centering
\caption{\textbf{Inference and End-to-End (E2E) Latency Analysis.} We compare compute runtime alongside a theoretical E2E latency model ($T_{E2E} = T_{compute} + N_{trips} \times 10$ ms). Interactive methods require multiple network trips, resulting in higher E2E latency than one-shot broadcast methods.}
\label{tab:latency}
\small 
\setlength{\tabcolsep}{5pt} 
\begin{tabular}{llccc}
\toprule
Method & Mode & Compute & Trips & Est. E2E \\
\midrule
CoBEVT (Dense) & Broadcast & 17 ms & 1 & 27 ms \\
CoBEVT + VQ & Broadcast & 21 ms & 1 & 31 ms \\
CoSDH \cite{xu_cosdh_2025} & Interactive & 35 ms & 2 & 55 ms \\
InfoCom \cite{infocom} & Broadcast & 37 ms & 1 & 47 ms \\
\textbf{Ours (Dual-Res)} & \textbf{Broadcast} & \textbf{40 ms} & \textbf{1} & \textbf{50 ms} \\
\bottomrule
\end{tabular}
\end{table}

\noindent \textbf{Analysis.}
As shown in Table~\ref{tab:latency}, standard CoBEVT is computationally the fastest (17 ms) but requires untenable bandwidth. State-of-the-art efficient methods like CoSDH~\cite{xu_cosdh_2025} achieve a low inference latency of 35 ms. CoSDH theoretically mitigates its interactive communication delays by parallelizing local backbone compute with the transmission of demand masks. However, this assumes ideal, uninterrupted channel access and implies that transmission delays scale strictly with data volume. 

To provide a fair end-to-end (E2E) system comparison, we introduce a standard theoretical network model: $T_{E2E} = T_{compute} + N_{trips} \times T_{hop}$, assuming a typical V2X transmission delay of $T_{hop} = 10$ ms.  In the kilobyte regime (e.g., $<2.0$ KB), transmission latency in IEEE 802.11-based V2X systems is not dominated by payload serialization time but by channel access delay and fixed PHY overheads. In CSMA/CA-based systems, packet transmission requires arbitration, random backoff, and inter-frame spacing, whose duration depends on contention rather than payload size~\cite{ieee80211,hartenstein_vanet}. Furthermore, each transmission incurs a fixed physical-layer preamble and PLCP header overhead that becomes dominant for small packets~\cite{chen_overhead_2012}. As a result, reducing payload size below the kilobyte level yields diminishing returns in latency, while multi-round interactive protocols incur the fixed channel-access cost multiple times per frame ($N_{trips} \ge 2$). In contrast, sender-driven broadcast methods like InfoCom~\cite{infocom} and our approach bypass these handshakes entirely ($N_{trips} = 1$).

\noindent \textbf{Real-World Implication.}
Our proposed method has a compute latency of \textbf{40 ms} (+5 ms vs CoSDH, +3 ms vs InfoCom). Crucially, because our approach operates in a strict \textbf{one-shot broadcast} mode, it transmits both the deterministic coarse base layer and prioritized fine patches in a single, kilobyte-scale payload, entirely avoiding multi-round handshake vulnerabilities. Under our E2E model, our total estimated latency is 50 ms, which is faster than the interactive CoSDH pipeline (55 ms) when accounting for unavoidable MAC-layer contention. Under this modeled communication setting, the estimated end-to-end latency remains below the 100 ms (10 Hz) operating interval, providing state-of-the-art deterministic coverage while maintaining a strict real-time advantage over interactive protocols.
\begin{figure}[h!]
    \centering
    \includegraphics[width=0.6\columnwidth]{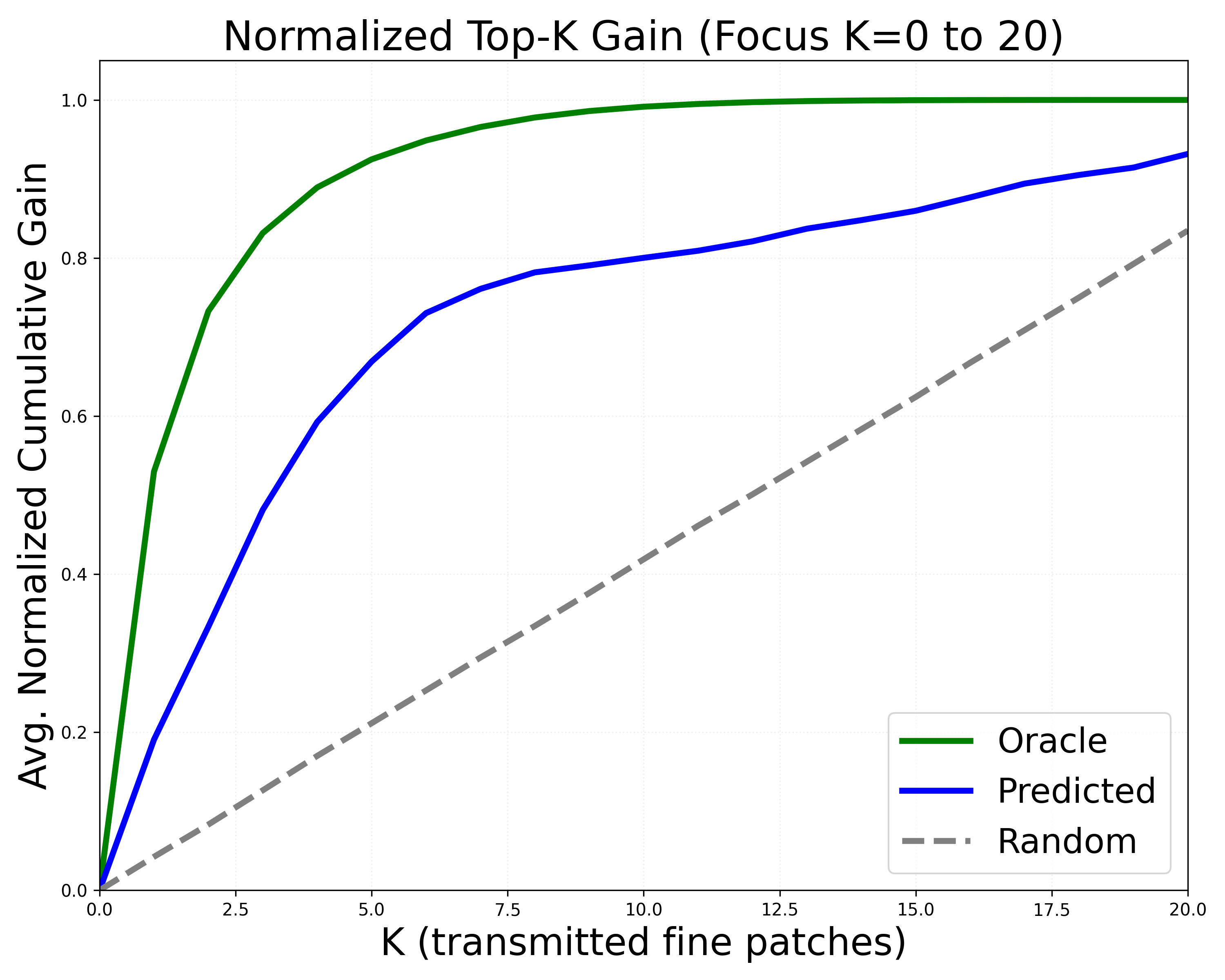}
    \caption{Per-agent Top-$K$ normalized cumulative marginal detection gain ($K \le 20$). Ranking patches by our predicted benefit score $\beta$ successfully concentrates the majority of attainable detection gain within a strict low-$K$ budget cap regime, vastly outperforming random allocation.}
    \label{fig:topk_gain}
\end{figure}
\section{Correlation of Benefit Score with Marginal Detection Gain}
\label{supp:utility_analysis}

To rigorously verify that our learned selection policy accurately prioritizes critical spatial regions, we measure the true marginal detection impact of refining a single spatial patch $i$. We define this marginal gain as the drop in detection loss when swapping the coarse reconstruction of a selected cell for a fine patch: $\Delta L_i = L_{det}(F_{coarse}) - L_{det}(F_{coarse \rightarrow fine(i)})$. 

By isolating the non-negative gain as $\Delta L_i^+ = \max(\Delta L_i, 0)$, we sort the available patches per agent by our network's predicted benefit score $\beta$ and compare it against an oracle ranking that uses the true $\Delta L_i^+$.
As illustrated in Fig.~\ref{fig:topk_gain}, prioritizing patches by $\beta$ captures a substantial fraction of the total attainable detection gain even under strict bandwidth constraints. Furthermore, as a dataset-level diagnostic, $\beta$ achieves a ROC-AUC of 0.785 for successfully identifying the highest-utility patches (defined as the top 15\% by $|\Delta L|$). This yields a Precision@10\% of 0.509 compared to a 0.10 random baseline, confirming that our predicted benefit score is strongly correlated with actual downstream task utility.
\section{Qualitative Analysis}
\begin{figure}[h!]
    \centering

    \begin{subfigure}[b]{0.49\textwidth}
        \centering
        \includegraphics[width=\linewidth]{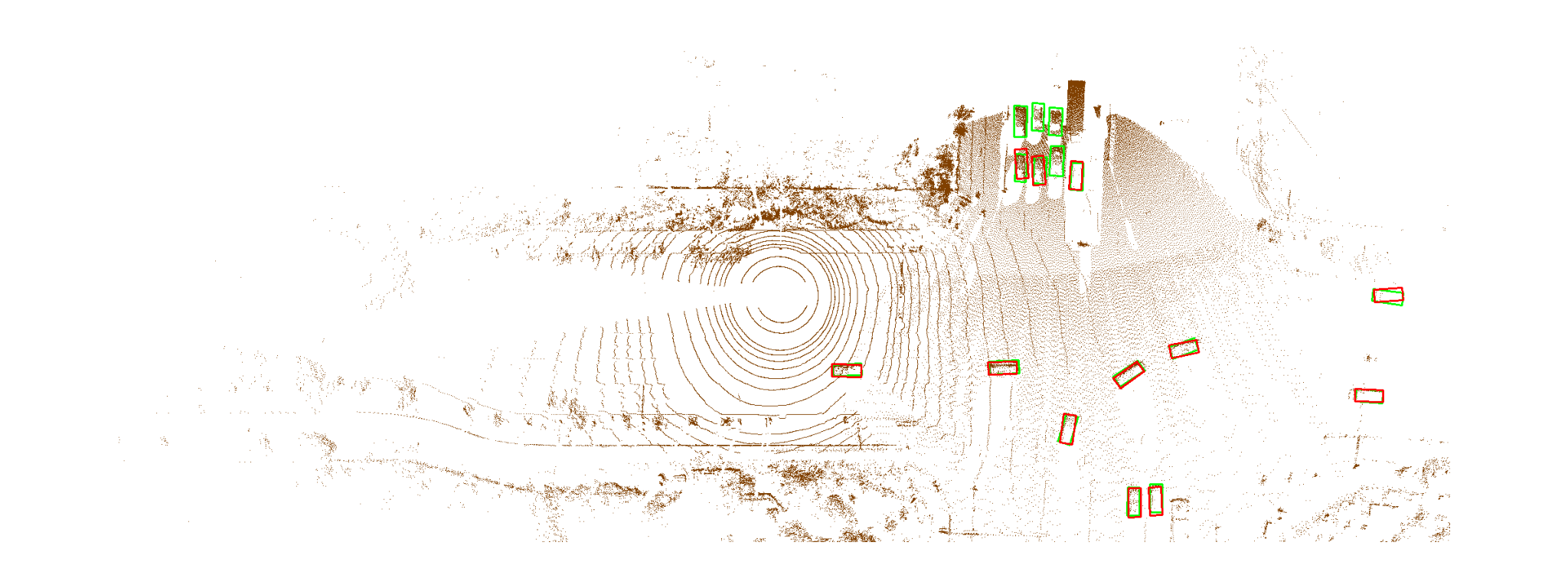}
        \caption{InfoCom}
        \label{fig:qual_infocom}
    \end{subfigure}
    \hfill
    \begin{subfigure}[b]{0.49\textwidth}
        \centering
        \includegraphics[width=\linewidth]{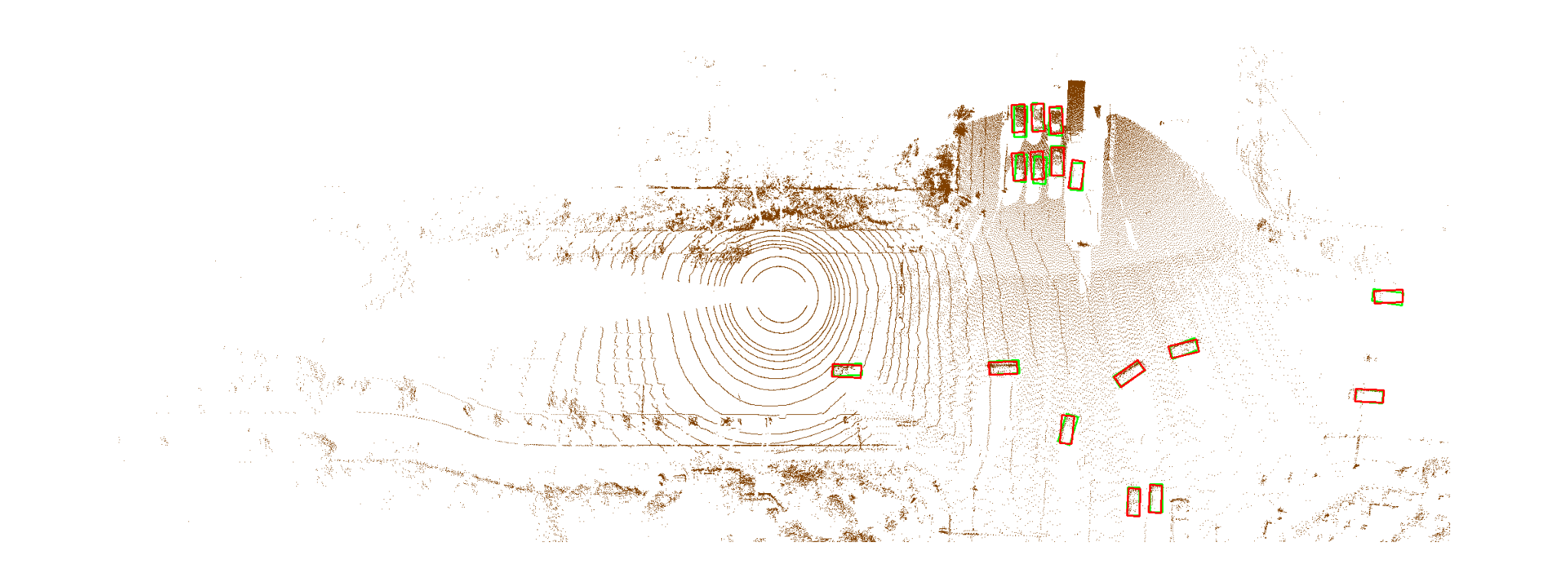}
        \caption{CoBEVT-DR (Ours)}
        \label{fig:qual_ours}
    \end{subfigure}

    \caption{\textbf{Qualitative comparison.}
    Green: ground truth. Red: detections.
    \textbf{(a)} InfoCom relies on pure sparsification, discarding unselected regions and missing several vehicles in the upper-right cluster.
    \textbf{(b)} CoBEVT-DR at a strict 2~KB budget recovers the cluster; the dense coarse floor provides a continuous semantic anchor even for unselected objects.}
    \label{fig:qualitative_comparison}
\end{figure}

Fig.~\ref{fig:qualitative_comparison} provides representative qualitative comparisons between InfoCom and CoBEVT-DR under the same scenes and payload regime. InfoCom communicates sparse task-relevant regions, but the missing spatial context can lead to incomplete object support or less stable predictions. In contrast, CoBEVT-DR preserves a dense coarse BEV floor and refines selected high-utility regions, maintaining scene-level context while recovering fine object details. These examples illustrate the main advantage of the proposed coverage-refinement design: sparse high-resolution communication without breaking the dense BEV structure expected by the fusion module.

\section{Task transfer to dynamic-object segmentation.}
As an additional task-transfer check, we evaluate CoBEVT-DR on dynamic-object BEV segmentation using the original CoBEVT dynamic-object segmentation setup. The segmentation BEV range is \(x,y\in[-50,50]\,\mathrm{m}\), with height range \(z\in[-3,1]\,\mathrm{m}\), following CoBEVT~\cite{xu_cobevt_2022}. The communication design, coarse/fine tokenization, and payload accounting remain unchanged; only the downstream objective and metric differ.

\begin{table}[h!]
    \centering
    \caption{\textbf{Task transfer to dynamic-object BEV segmentation.}
    We evaluate the same coverage-refinement communication design on dynamic-object segmentation using the original CoBEVT segmentation setup. Payload is reported per non-ego agent.}
    \label{tab:task_transfer_dynamic_segmentation}
    \scriptsize
    \setlength{\tabcolsep}{6pt}
    \begin{tabular}{l c c}
        \toprule
        Method & Dynamic IoU$\uparrow$ & KB$\downarrow$ \\
        \midrule
        CoBEVT & 0.47 & 524.28 \\
        CoBEVT-SimVQ & 0.43 & 0.72 \\
        \midrule
        CoBEVT-DR & 0.47 & 0.40 \\
        CoBEVT-DR & 0.46 & 0.20 \\
        CoBEVT-DR & 0.37 & 0.09 \\
        \bottomrule
    \end{tabular}
\end{table}

As shown in~\cref{tab:task_transfer_dynamic_segmentation}, CoBEVT-DR reaches dense-level dynamic-object IoU at 0.40~KB and remains close to the dense baseline at 0.20~KB per non-ego agent, while using orders of magnitude less communication than dense CoBEVT. At the more aggressive 0.09~KB operating point, performance decreases but still provides a useful low-bandwidth fallback. Compared with uniform SimVQ compression, coverage-refinement achieves higher IoU at substantially lower payload, suggesting that the proposed allocation strategy transfers beyond 3D object detection under strict byte budgets.
\section{Additional Evaluation on V2X-Real}
\label{sec:supp_v2xreal}

We additionally evaluate our model on the real-world multi-class
V2X-Real-VC benchmark~\cite{xiang2024v2x} under a strict
2.0\,KB per-agent payload cap. As shown in Table~\ref{tab:v2xreal_comparison},
our model reaches 66.8/56.8 mAP at IoU thresholds 0.3/0.5,
with 89.3/86.6 AP for vehicles, 47.9/24.1 AP for pedestrians,
and 63.1/59.8 AP for trucks. Published results use different
detector architectures and evaluation implementations and are
therefore included as contextual rather than strictly controlled
comparisons.

\begin{table}[h!]
    \centering
    \caption{
    Multi-class detection performance on the V2X-Real-VC test set.
    Car, Ped., and Truck report AP@0.3/AP@0.5.
    Published results are taken from the respective papers and may use
    different detector architectures and evaluation implementations.}
    \label{tab:v2xreal_comparison}
    \scriptsize
    \setlength{\tabcolsep}{1.8pt}
    \renewcommand{\arraystretch}{1.05}
    \resizebox{\columnwidth}{!}{
    \begin{tabular}{lccccc}
        \toprule
        Method &
        \textbf{Car AP@0.3/0.5}$\uparrow$ &
        \textbf{Ped. AP@0.3/0.5}$\uparrow$ &
        \textbf{Truck AP@0.3/0.5}$\uparrow$ &
        \textbf{mAP@0.3}$\uparrow$ &
        \textbf{mAP@0.5}$\uparrow$ \\
        \midrule

        No Fusion~\cite{xiang2024v2x}
        & 38.7/35.9 & 25.5/13.1 & 20.2/14.5 & 28.2 & 21.2 \\

        Early Fusion~\cite{xiang2024v2x}
        & 51.1/47.6 & 31.6/16.0 & 32.5/23.6 & 38.4 & 29.1 \\

        F-Cooper~\cite{xiang2024v2x}
        & 57.3/54.2 & 30.0/14.1 & 27.0/21.2 & 38.1 & 29.8 \\

        AttFuse~\cite{xiang2024v2x}
        & 62.6/59.4 & 32.2/15.5 & 32.6/26.6 & 42.5 & 33.8 \\

        V2X-ViT~\cite{xiang2024v2x}
        & 62.7/60.3 & 36.7/18.6 & 35.1/28.3 & 44.8 & 35.8 \\

        CooPre~\cite{zhao2025coopre}
        & 71.5/70.2 & 46.9/28.0 & 61.9/58.3 & 60.1 & 52.2 \\

        FocalComm~\cite{shenkut2026focalcomm}
        & 91.5/89.6 & 57.4/27.3 & 53.9/51.6 & 67.6 & 56.1 \\

        \midrule

        \textbf{CoBEVT-DR} (2 KB)
        & \textbf{89.3/86.6}
        & \textbf{47.9/24.1}
        & \textbf{63.1/59.8}
        & \textbf{66.8}
        & \textbf{56.8} \\

        \bottomrule
    \end{tabular}}
\end{table}

\section{Additional Evaluation on V2XVerse}
\label{sec:v2xverse}

To evaluate whether the proposed byte-constrained representation transfers
beyond the perception benchmarks used in the main paper, we additionally
evaluate it on V2XVerse~\cite{codriving}. We report both multi-class object detection and
open-loop ego-planning performance under increasing communication budgets.
The communication mechanism is unchanged: each collaborator transmits the
coarse base representation together with budgeted high-resolution refinement
patches.

Table~\ref{tab:v2xverse_perception} compares our method against representative
V2XVerse fusion baselines. Even under kilobyte-scale communication, the
proposed representation achieves strong detection performance across all
three object classes. At only 2\,KB per collaborator, our method reaches
73.20 mAP@0.3, compared with 68.0 for CoDriving. Increasing the budget to
3\,KB further improves mAP@0.3 to 74.67.

The gains are particularly pronounced for vulnerable road users. From the
coarse-only representation to 3\,KB, AP@0.5 improves by 4.71 points for
vehicles, 9.76 points for pedestrians, and 14.47 points for bicyclists.
This indicates that sparse high-resolution refinement is especially beneficial
for more challenging object classes, rather than merely improving already
strong vehicle detections.

\begin{table}[t]
\centering
\caption{Multi-class detection performance on V2XVerse. Our DR variants
operate under strict per-agent communication budgets. Values are AP. The results are taken from CoDriving~\cite{codriving}.}
\label{tab:v2xverse_perception}
\resizebox{\columnwidth}{!}{%
\begin{tabular}{lcccccccc}
\toprule
\multirow{2}{*}{\textbf{Method}}
& \multicolumn{3}{c}{\textbf{Vehicle}}
& \multicolumn{2}{c}{\textbf{Bicyclist}}
& \multicolumn{2}{c}{\textbf{Pedestrian}}
& \multirow{2}{*}{\textbf{mAP30}} \\
\cmidrule(lr){2-4}
\cmidrule(lr){5-6}
\cmidrule(lr){7-8}
& \textbf{AP30} & \textbf{AP50} & \textbf{AP70}
& \textbf{AP30} & \textbf{AP50}
& \textbf{AP30} & \textbf{AP50}
& \\
\midrule
No Fusion
& 0.89 & 0.84 & 0.73
& 0.40 & 0.30
& 0.41 & 0.24
& 0.57 \\

Late Fusion
& 0.88 & 0.86 & 0.81
& 0.43 & 0.38
& 0.45 & 0.27
& 0.59 \\

F-Cooper
& 0.93 & 0.82 & 0.68
& 0.44 & 0.29
& 0.56 & 0.33
& 0.64 \\

V2X-ViT
& 0.93 & 0.91 & 0.84
& 0.50 & 0.36
& 0.41 & 0.12
& 0.61 \\

CoopDet3D
& 0.93 & 0.90 & 0.81
& 0.48 & 0.41
& 0.53 & 0.31
& 0.65 \\

CoDriving
& 0.94 & 0.91 & 0.83
& 0.52 & 0.41
& 0.58 & 0.35
& 0.68 \\

\midrule
CoBEVT-DR -- Coarse ($\sim$330 B)
& 0.9164 & 0.9054 & 0.8545
& 0.5037 & 0.4097
& 0.5246 & 0.2874
& 0.6482 \\

CoBEVT-DR -- 1 KB
& 0.9106 & 0.9025 & 0.8685
& 0.5235 & 0.4561
& 0.5646 & 0.3379
& 0.6662 \\

CoBEVT-DR -- 2 KB
& 0.9515 & 0.9396 & 0.9034
& 0.6103 & 0.5372
& 0.6341 & 0.3779
& 0.7320 \\

CoBEVT-DR -- 3 KB
& \textbf{0.9648} & \textbf{0.9525} & \textbf{0.9125}
& \textbf{0.6282} & \textbf{0.5544}
& \textbf{0.6472} & \textbf{0.3850}
& \textbf{0.7467} \\
\bottomrule
\end{tabular}}
\end{table}

{
    \small
    \bibliographystyle{ieeenat_fullname}
    \bibliography{main}
}

\end{document}